\documentclass[10pt,journal,compsoc]{IEEEtran}

\usepackage{caption}
\usepackage{times}
\usepackage{graphicx}
\usepackage{amsmath}
\usepackage{amssymb}
\usepackage{booktabs}
\usepackage{makecell}
\usepackage{multirow}
\usepackage{threeparttable}
\usepackage{colortbl}
\usepackage{tabu}
\usepackage{xspace}
\usepackage{pifont}
\usepackage{enumitem}
\newcommand{\cmark}{\ding{51}\xspace}%
\newcommand{\xmarkg}{\textcolor{lightgray}{\ding{55}}\xspace}%

\usepackage{amssymb,graphicx}
\newcommand\vcent[1]{\vcenter{\hbox{#1}}}
\newcommand\loudspeaker[1][3]{\ensuremath{\vcent{\rule{.6ex}{.6ex}}\kern-.5ex%
  \vcent{\scalebox{.6}[1]{\rotatebox[origin=center]{90}{$\blacktriangle$}}}%
  \ifnum#1>0\relax\kern.1ex\vcent{\scalebox{.3}{)}}\ifnum#1>1\relax\kern-.15ex%
  \vcent{\scalebox{.4}{)}}\ifnum#1>2\relax\kern-.23ex\vcent{\scalebox{.5}{)}}%
  \fi\fi\fi}%
}

\usepackage{epsfig}
\newcommand{\pub}[1]{\color{gray}{\scriptsize{[{#1}]}}}

\usepackage{mathrsfs}
\usepackage{soul}
\newcommand{\ie}{{\emph{i.e.}}\xspace}
\newcommand{\eg}{{\emph{e.g.}}\xspace}

\newcommand{\vs}{{\emph{vs.}}\xspace}
\newcommand{\etal}{{\emph{et al.}}\xspace}
\newcommand{\etc}{{\emph{etc.}}\xspace}
\usepackage{array}
\usepackage{tabu}

\usepackage[pagebackref=true,breaklinks,colorlinks]{hyperref}

\usepackage[capitalize]{cleveref}
\crefname{section}{Sec.}{Secs.}
\Crefname{section}{Section}{Sections}
\Crefname{table}{Table}{Tables}
\crefname{table}{TABLE}{TABLE}

\newcommand{\Rmnum}[1]{\expandafter\@slowromancap\romannumeral #1@}

\usepackage{color}
\usepackage{xcolor}
\newcommand{\numvideo}{2,006\xspace}
\newcommand{\numobject}{8,171\xspace}

\newcommand{\numsentence}{33,072\xspace}
\newcommand{\ourdataset}{MeViS\xspace}
\newcommand{\ourdatasetICCV}{MeViSv1\xspace}
\newcommand{\ourdatasetTPAMI}{MeViSv2\xspace}

\newcommand{\fullname}{\textbf{M}otion \textbf{e}xpressions \textbf{Vi}deo \textbf{S}egmentation\xspace}

\definecolor{mycolor}{RGB}{0, 0, 255}

\newcommand{\ourmodel}{LMPM\xspace}
\newcommand{\ournewmodel}{LMPM++\xspace}
\usepackage{pifont}
\newcommand{\myparagraph}[1]{{\vspace{.3em} \noindent \bf #1}}

\let\oldsubsection\subsection
\renewcommand{\subsection}[1]{\oldsubsection{#1} }
\ifCLASSOPTIONcompsoc
  \usepackage[nocompress]{cite}
\else
  \usepackage{cite}
\fi
\ifCLASSINFOpdf
\else
\fi
\begin{document}
%
\title{MeViS: A Multi-Modal Dataset for Referring Motion Expression Video Segmentation}
%
%
%
%

\author{Henghui~Ding,
        Chang~Liu,
        Shuting~He,
        Kaining~Ying,
        Xudong~Jiang,~\IEEEmembership{Fellow,~IEEE},\\
        Chen Change Loy,~\IEEEmembership{Senior~Member,~IEEE},
        Yu-Gang Jiang,~\IEEEmembership{Fellow,~IEEE}
\IEEEcompsocitemizethanks{
\IEEEcompsocthanksitem Henghui Ding, Kaining Ying, and Yu-Gang Jiang are with Fudan University, China 200433. (e-mail: henghui.ding@gmail.com)
\IEEEcompsocthanksitem Chang Liu and Shuting He are with Shanghai University of Finance and Economics, Shanghai, China, 200433.
\IEEEcompsocthanksitem Xudong Jiang and Chen Change Loy are with Nanyang Technological University, Singapore 639798.
}
}

%
%

\markboth{IEEE TRANSACTIONS ON PATTERN ANALYSIS AND MACHINE INTELLIGENCE}%
{Shell \MakeLowercase{\textit{et al.}}: Bare Demo of IEEEtran.cls for Computer Society Journals}
\IEEEtitleabstractindextext{
\begin{abstract}
   This paper proposes a large-scale multi-modal dataset for referring motion expression video segmentation, focusing on segmenting and tracking target objects in videos based on language description of objects' motions. Existing referring video segmentation datasets often focus on salient objects and use language expressions rich in static attributes, potentially allowing the target object to be identified in a single frame. Such datasets underemphasize the role of motion in both videos and languages. To explore the feasibility of using motion expressions and motion reasoning clues for pixel-level video understanding, we introduce MeViS, a dataset containing 33,072 human-annotated motion expressions in both text and audio, covering 8,171 objects in 2,006 videos of complex scenarios. We benchmark 15 existing methods across 4 tasks supported by MeViS, including 6 referring video object segmentation (RVOS) methods, 3 audio-guided video object segmentation (AVOS) methods, 2 referring multi-object tracking (RMOT) methods, and 4 video captioning methods for the newly introduced referring motion expression generation (RMEG) task. The results demonstrate weaknesses and limitations of existing methods in addressing motion expression-guided video understanding. We further analyze the challenges and propose an approach LMPM++ for RVOS/AVOS/RMOT that achieves new state-of-the-art results. Our dataset provides a platform that facilitates the development of motion expression-guided video understanding algorithms in complex video scenes. The proposed MeViS dataset and the method's source code are publicly available at \href{https://henghuiding.com/MeViS/}{https://henghuiding.com/MeViS/}.
\end{abstract}

}

\maketitle


\IEEEdisplaynontitleabstractindextext

%
\IEEEpeerreviewmaketitle

\IEEEraisesectionheading{\section{Introduction}\label{sec:intro}}
\IEEEPARstart{R}{eferring} video segmentation is an emerging field that aims at segmenting and tracking the specific target object referred by a given natural language expression~\cite{MeViS,ReferringSegSurvey,khoreva2018video,seo2020urvos}. This task has traditionally been a subset of semi-supervised video object segmentation, where the clue of target object is provided through means such as a mask, scribble, or sentence in the first frame. Existing datasets in this context, such as DAVIS$_{16}$-RVOS~\cite{khoreva2018video} and Refer-YouTube-VOS~\cite{seo2020urvos}, typically encompass videos featuring isolated and salient objects with evident static characteristics. The corresponding expressions frequently contain static attributes like the object's color and shape, which can be identified from a single frame. Consequently, motion properties of videos are less pronounced in these expressions, and methods designed for referring image segmentation can effectively be applied to referring video segmentation, yielding favorable performance \cite{khoreva2018video,vltpami,bellver2020refvos,liu2021cmpc}.

\begin{figure*}
    \centering
    \includegraphics[width=0.999\textwidth]{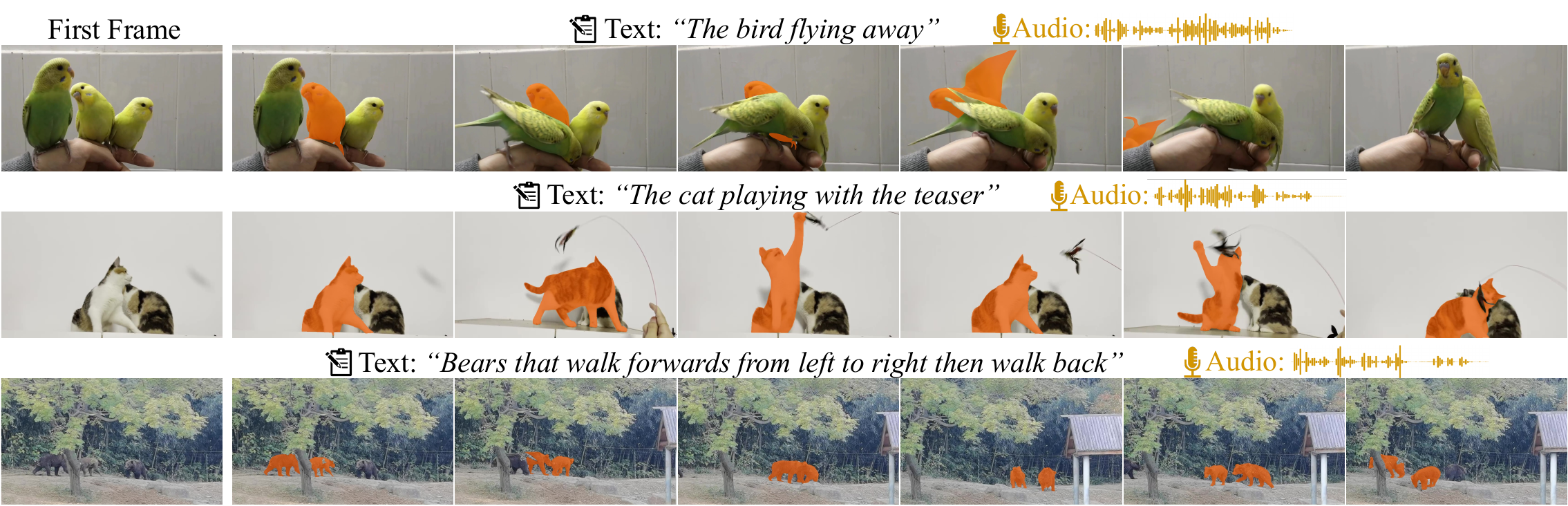}
    \vspace{-6mm}
\caption{ Examples from \fullname (\textbf{\ourdataset}) showing the dataset's nature and complexity. The expressions in \ourdataset primarily focus on motion attributes, making it impossible to identify the target object from a single frame.  For example, the first example has three parrots with similar appearances, and the target object is identified as \textit{``The bird flying away''}. This object can only be recognized by capturing its motion throughout the video. The updated \ourdatasetTPAMI further provides motion-reasoning and no-target expressions (see \cref{fig:motionreasoning}), adds audio \loudspeaker~expressions alongside text, and provides mask and bounding box trajectory annotations.
}\label{Fig:teaser}
\end{figure*}

The motivation of this work is to emphasize the importance of temporal motion characteristics in videos and explore the feasibility of employing motion-related expressions to identify and segment objects within video content. To this end, we propose a new large-scale dataset named \textbf{M}otion \textbf{e}xpressions \textbf{Vi}deo \textbf{S}egmentation (\textbf{\ourdataset}). Some samples of MeViS are shown in \cref{Fig:teaser}. The \ourdataset dataset contains \numvideo videos with a total of \numobject distinct objects. In the conference version, \ourdatasetICCV~\cite{MeViS}, 28,570 motion-related expressions are provided for referring and delineating these objects, focusing on direct motion descriptions of single or multiple targets. Compared to \ourdatasetICCV~\cite{MeViS}, the updated \ourdatasetTPAMI in this work significantly expands the dataset with more challenging motion expressions, adding audio format expressions, providing tracking annotations, and supporting more tasks. First, the updated dataset includes 4,502 new challenging expressions, bringing the total to \numsentence expressions—the largest in the field of referring video. These additions include motion reasoning expressions, which involve implicit queries requiring complex reasoning, and no-target expressions, which are deceptive motion descriptions that relate to the video but do not refer to any actual object, as shown in \cref{fig:motionreasoning}. In addition to text expressions, the updated \ourdatasetTPAMI further provides above 150,000 seconds audio expressions, facilitating the study of audio-guided video object segmentation (AVOS) and multi-modal referring expressions. Furthermore, we provide tracking annotations in \ourdatasetTPAMI, establishing it as the largest referring multi-object tracking (RMOT) dataset. Beyond perception tasks, we introduce a new task based on MeViS: Referring Motion Expression Generation (RMEG). This task aims to generate an unambiguous and concise motion expression for the selected objects in a given video.

In the construction of the \ourdataset dataset, several steps are undertaken to highlight the temporal motions inherent to videos. First, an assortment of videos is selected with the criterion that they showcase multiple interacting objects in motion, deliberately excluding low-quality videos where isolated objects could be easily described through static attributes alone. Second, the dataset prioritizes language expressions that focus on motion clues (\eg, walking, moving) rather than static clues (\eg, color, shape). These rules distinguish our \ourdataset from earlier datasets like~\cite{gavrilyuk2018actor,khoreva2018video,seo2020urvos}, which contain salient targets in their videos or include obvious static clues in their sentence annotations. \ourdataset also sets itself apart from referring image segmentation datasets, such as~\cite{yu2016modeling,mao2016generation,wu2020phrasecut,kazemzadeh-etal-2014-referitgame,GRES}, by considering the temporal properties of video, which is overlooked in these datasets. Uniquely, unlike existing referring video segmentation datasets~\cite{gavrilyuk2018actor,khoreva2018video,seo2020urvos} that are limited to single-object expressions, \ie, one expression refers to only one target object, \ourdataset broadens the scope to further support generalized expressions that can refer to an unlimited number of target objects, including no target, thereby enhancing the generalizability and real-world applicability of the \ourdataset dataset and referring video segmentation.

The proposed \ourdataset dataset presents significant challenges in capturing and understanding motion in both video and language. Language expressions may refer to actions spanning varying numbers of frames, necessitating the capture of both fleeting movements and long-term actions throughout the entire video. This requirement introduces substantial challenges in comprehending motion within the video content and the corresponding language expressions. Capturing fleeting movements necessitates detailed attention to individual frames, whereas understanding long and complex movements requires maintaining temporal context across the entire video. To evaluate the effectiveness of existing methods in addressing the challenges posed by \ourdatasetTPAMI, we benchmark 15 existing methods across 4 tasks and conduct comprehensive comparisons, including 6 referring video object segmentation (RVOS) methods~\cite{vltpami,wu2022referformer,MTTR,Ding_2022_CVPR,seo2020urvos,DsHmp}, 3 Audio-guided Video Object Segmentation (AVOS) methods~\cite{pan2022wnet,MeViS,mutr}, 2 Referring Multi-Object Tracking (RMOT) methods~\cite{rmot,zhang2024bootstrapping_rmot}, and 4 video captioning methods~\cite{wang2022git,chen2024vast,nadeem2024narrativebridge,cheng2024videollama2advancingspatialtemporal}. The experimental results show that \ourdataset presents greater challenges than existing datasets, revealing that existing methods are insufficient in effectively addressing motion expression-related video understanding.

In addition to introducing the \ourdataset dataset, we propose a baseline method: Language-guided Motion Perception and Matching (\ournewmodel). \ournewmodel utilizes language-conditional queries to detect potential target objects within the video, and represents these objects with object embeddings. In this way, it enhances robustness and computational efficiency compared to using frame features~\cite{VITA}. We then feed object embeddings into a large language model (LLM), capturing and reasoning temporal context and achieving a comprehensive understanding of the video. 
Unlike existing methods~\cite{TrackGPT,VISA}, we generate and input object tokens instead of frame features into LLM, enabling it to process much longer sequences, \eg, 200 frames compared to 3~\cite{TrackGPT} or 13 frames~\cite{VISA} in previous methods. Understanding the sequence of movements is crucial. For example, the actions of ``\textit{first jumping high and then jumping far}'' \vs ``\textit{first jumping far and then jumping high}'', though similar in overall words, represent distinct motion patterns. To address this, we introduce a temporal-level contrastive loss, enabling the model to differentiate motions with different temporal orders or sequences.

In a summary, our main contributions are as follows:
\begin{itemize}
\setlength\itemsep{0.1em}
    \item We build \textbf{\ourdatasetTPAMI}, a large-scale multi-modal referring motion expression video segmentation dataset focusing on segmenting and tracking object(s) in the given video indicated by a motion expression in either text or audio format.
    \item The proposed \ourdatasetTPAMI dataset can support at least 4 different referring video tasks: referring video object segmentation (RVOS), audio-guided video object segmentation (AVOS), referring multi-object tracking (RMOT), and referring motion expression generation (RMEG).
    \item We benchmark 15 methods across RVOS, AVOS, RMOT, and RMEG tasks on the proposed \ourdatasetTPAMI dataset, serving as a reference for future works in these 4 tasks on \ourdatasetTPAMI.
    \item Taking a close look at the proposed \ourdataset dataset, we identify several challenges and develop a baseline approach for perception tasks, named Language-guided Motion Perception and Matching (\ournewmodel), to meet these challenges.
    \item We discuss potential directions for future video-language motion understanding research. 
\end{itemize}

\section{Related Work}
\subsection{Referring Image Segmentation}
Referring image segmentation~\cite{ding2021vision,vltpami,GRES} aims at segmenting the target object in the given image referred by a natural language expression describing the target's properties, \eg, location and color. 
Since introduced by Hu~\etal~in 2016~\cite{hu2016segmentation}, this task has attracted significant interest and attention. Before the advent of Transformer-based models, conventional methods~\cite{li2018referring,liu2017recurrent} commonly relied on Fully Convolutional Networks (FCN)~\cite{long2015fully,ding2018context} and Recurrent Neural Networks (RNN) to extract image and language features, respectively. Subsequently, these multi-modal features were integrated using specifically designed modules. For example, Liu \etal~\cite{liu2017recurrent} present the Recurrent Multimodal Interaction (RMI) module to iteratively merge the features of individual words into the image features. 
Apart from one-stage methods, there are methods that decompose the task into two stages: instance segmentation and language-object matching~\cite{yu2018mattnet, ISFP, jing2021locate}.
For example, Yu \etal employ the pre-trained instance segmentation model Mask R-CNN~\cite{he2017mask} to detect all instances within an image. Then, they select the instance that best matches the given expression as the final output.
Contextual modeling of both language and visual information is essential, and numerous studies have explored this direction~\cite{hui2020linguistic,yang2021bottom,ye2019cross}.
As an example, Ye \etal introduce Cross-Modal Self-Attention (CMSA)~\cite{ye2019cross} to identify the most relevant words within the language expression and pixels within the image, enhancing contextual comprehension.

Recently, the impressive achievements of Transformer~\cite{vaswani2017attention} in various vision-related tasks have inspired many works in referring image segmentation~\cite{OpenVocSurvey}.
Ding \etal~\cite{vltpami,ding2021vision} are the pioneers in introducing Transformer into referring segmentation and introduce a Vision-Language Transformer (VLT).
Following Ding \etal~\cite{vltpami,ding2021vision}, more Transformer-based methods have emerged in the field~\cite{yang2021lavt, wang2022cris, kim2022restr,CGFormer,PolyFormer,GRES,UNINEXT}. For example, Wang \etal~\cite{wang2022cris} present a Vision-Language Decoder to handle visual and text tokens extracted using CLIP~\cite{radford2021learning}. Yang \etal~\cite{yang2021lavt} focus on the fusion of multi-modal features and introduce a Language-Aware Vision Transformer (LAVT). These advancements showcase the growing influence of Transformer models in this area.

\subsection{Referring Video Segmentation} 

Referring video object segmentation is an emerging multi-modal video understanding task~\cite{wang2020context,ningpolar,wang2019asymmetric,mcintosh2020visual,liu2021cmpc,hui2021collaborative,Wu_2022_CVPR,zhao2022modeling,sun2022starting,chen2022multi,yang2022tubedetr,tang2021human} that focuses on segmenting the target object specified by a given expression throughout a video. It is first introduced in 2018 by A2D~\cite{gavrilyuk2018actor} and DAVIS$_{17}$-RVOS~\cite{khoreva2018video}. The A2D dataset~\cite{gavrilyuk2018actor} aims to segment actors based on descriptions of their actions within video content, whereas DAVIS$_{17}$-RVOS~\cite{khoreva2018video} utilizes language, rather than masks, as the reference for the target object in video object segmentation. Subsequently, Seo \etal~\cite{seo2020urvos} developed Refer-YouTube-VOS, which is based on the YouTube-VOS-2019 dataset~\cite{xu2018youtube}. These datasets usually provide expressions rich in static attributes describing a single object. To emphasize motion, Ding \etal~\cite{MeViS} introduce MeViS dataset with numerous motion expressions.

Existing methods typically treat referring video segmentation as a variant of semi-supervised video object segmentation~\cite{davis2017} by replacing mask references with language references. For example, Khoreva \etal~\cite{khoreva2018video} adapt the referring image segmentation method MAttNet~\cite{yu2018mattnet} to achieve frame-level segmentation, followed by post-processing to ensure temporal consistency. URVOS~\cite{seo2020urvos} utilizes cross-modal attention to perform per-frame segmentation, propagating the mask across frames using a memory attention module. RefVOS~\cite{bellver2020refvos} segments each frame independently based on fused language and image/frame features, without leveraging temporal information. Liang \etal~\cite{liang2021topdown} propose a top-down approach that first detects all object tracklets and then selects the target object by matching language and tracklet features. More recently, ReferFormer~\cite{wu2022referformer}, MTTR~\cite{MTTR}, and DsHmp~\cite{DsHmp} employ Transformers~\cite{vaswani2017attention} to address referring video object segmentation.

\subsection{Audio-guided Video Segmentation}
Audio-guided Video Segmentation introduces the audio modality into video segmentation, with three main settings: audio-visual segmentation (AVS) \cite{avsbench,avsbench-semantic}, audio-guided video object segmentation (AVOS) \cite{pan2022wnet}, and referring audio-visual segmentation (Ref-AVS) \cite{refavs}. The goal of AVS \cite{avsbench,avsbench-semantic} is to segment the sound-emitting objects in a video, such as birds chirping, cars honking, or dogs barking. In AVOS \cite{pan2022wnet}, the audio is human speech, and the goal is to segment the object described by the speech. The recently proposed Ref-AVS \cite{refavs} focuses on segmenting objects in videos based on referring expressions that consider both visual and audio signals, unlike RVOS, which considers visual and textual signals. \ourdatasetTPAMI focuses on AVOS, which has significant applications in embodied scenarios. The AVOS-Bench \cite{pan2022wnet} dataset, derived from existing RVOS datasets with text converted to speech through human narration, has noisy speech and limits model complexity. 
\ourdatasetTPAMI increases the challenge by using both TTS \cite{seedtts} and human narration for speech generation, inheriting the complexity of MeViS's videos and descriptions, which further enhances the difficulty and practicality of \ourdatasetTPAMI.

\subsection{Referring Object Tracking}
Different from the above mentioned referring segmentation tasks, referring object tracking \cite{shao_rsot, rmot} aims to detect the corresponding bounding box tracklets based on the language description. There are two common settings: referring single-object tracking (RSOT) \cite{rsot,shao_rsot,yang2020grounding_rsot,feng2020real_rsot} and referring multi-object tracking (RMOT) \cite{rmot,du2024ikun,zhang2024bootstrapping_rmot}.
RSOT, defined by Li~\etal\cite{rsot}, focuses on localizing a single target object in a video based on a sentence for the first frame. Yang~\etal\cite{yang2020grounding_rsot} and Feng~\etal\cite{feng2020real_rsot} divide this task into grounding and tracking. The grounding stage detects the language referred target, while the tracking stage tracks the target in subsequent based on the first-frame grounding result. \cite{yang2020grounding_rsot} also performs visual matching based on the history of grounded objects and language-based grounding for each frame.

Despite significant advancements in RSOT, current RSOT methods are constrained by their ability to describe only a single target per expression, and the language description is typically provided solely for the first frame. These limitations hinder their effectiveness in real-world applications.
To address these issues, Wu~\etal~\cite{rmot} propose RMOT and introduce the Refer-KITTI benchmark, designed to handle multi-object and temporally status-variant scenarios. Their baseline method extends the end-to-end multi-object tracking framework, MOTR \cite{zeng2022motr}, to accommodate cross-modal input. Subsequently, iKUN \cite{du2024ikun} adopts a two-stage approach, initially extracting object tracklets explicitly and then selecting those that correspond to the given language expression.
Later, Zhang \etal\cite{zhang2024bootstrapping_rmot} propose a query-based temporal enhanced framework, modeling long-term spatial-temporal interactions through transformer query features. Alongside this solution, they introduce a RMOT dataset, Refer-KITTI-V2, which includes diverse and multifaceted textual descriptions encompassing appearance, complex motion, position, \etc

MeViS supports more generalized referring understanding between expressions and long-term target states by including single-target, multi-target, and no-target expressions. Unlike previous RSOT and RMOT datasets, MeViS is designed for a broader range of scenarios and emphasizes finer-grained pixel-level perception, enhancing its applicability in real-world applications.

\section{\ourdataset Dataset}\label{sec:MeViS_Dataset}

\begin{figure}
    \centering
    \includegraphics[width=0.95\linewidth]{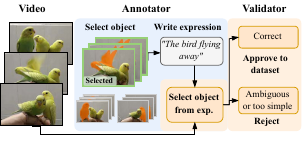}
    \vspace{-1.96mm}
    \caption{Flowchart of the language expression annotation and validation process of MeViS. Samples that are found ambiguous or too simple by validators will be rejected and discarded.}
    \label{fig:annotation}
\end{figure}

\subsection{Motion Expression Annotation}\label{sec:videoannotation}
\myparagraph{Video Collection.} We aim to build a challenging video dataset that includes a wide range of scenes to facilitate motion understanding. Based on publicly available video segmentation datasets with high-quality mask annotations~\cite{OVIS,UVO,TAOVOS,MOSE}, we select those that satisfy our criteria for motion and object complexity. The video selection process adheres to the following rules:
\begin{itemize}
\setlength\itemsep{0em}
\item[R1.] In \ourdataset, we select videos having multiple objects within the frame; videos containing only one or two salient objects are not considered. We particularly seek videos that contain objects with similar appearances, exemplified by the first video in \figurename~\ref{Fig:teaser}, which shows three similar looking parrots.

\item[R2.] We select videos containing objects that demonstrate substantial motion and movement. Videos with objects that display minimal or no movement are excluded from \ourdataset.
\end{itemize}

After reviewing over 4,000 candidate videos, we carefully selected the most appropriate and suitable videos that meet our requirements. By prioritizing quality over quantity, we finally chose \numvideo videos to create a dataset that is diverse and representative of a wide range of real-world complex video scenarios. 

The language expression annotation procedure for \ourdataset follows GRES~\cite{GRES} and ReferIt~\cite{kazemzadeh-etal-2014-referitgame}, using an interactive game-like approach that involves two players taking turns to annotate and validate. An overview of the language expression annotation and validation process is shown in \cref{fig:annotation}. The process of language expression annotation and validation is detailed as follows.


\begin{table*}[t]
\centering
\caption{Statistics of representative language-guided video segmentation datasets. The newly constructed \ourdatasetTPAMI has the largest number of objects and language expressions. More importantly, \ourdataset specifically focuses on segmenting objects in videos as indicated by motion expressions, supports generalized referring expressions for multi-target/no-target, and provide audio format. This dataset facilitates the exploration of using motion expressions for object segmentation and grounding in videos.}
\vspace{-2mm}
\begin{threeparttable}
\setlength\tabcolsep{5.5pt}
\renewcommand\arraystretch{1.3}
\begin{tabular}{lcccccccccccc}
\hline
\rowcolor[gray]{.9}Dataset~~~~~~~~~&Year &Pub. &Video&Object &Expression & Mask & \makecell[c]{Object/\\Video} &\makecell[c]{Target Object/\\Experission}  &\makecell[c]{Single-\\target} &\makecell[c]{Multi-\\target}&\makecell[c]{No-\\target} &Audio~\loudspeaker \\
\hline
\hline
A2D Sentence~\cite{gavrilyuk2018actor} &2018 &\pub{CVPR} &\makebox[5ex][r]{3,782}&\makebox[5ex][r]{4,825} &\ \ \makebox[6ex][r]{6,656} & 58k & 1.28 & 1&6,656 &\xmarkg&\xmarkg&\xmarkg\\
\rowcolor[gray]{.96}J-HMDB Sentence~\cite{gavrilyuk2018actor}&2018 &\pub{CVPR}  &\makebox[5ex][r]{928}&\makebox[5ex][r]{928} &\makebox[6ex][r]{928} &31.8k& 1 & 1 &\ \ \ \  928&\xmarkg&\xmarkg&\xmarkg\\
DAVIS$_{16}$-RVOS~\cite{khoreva2018video} &2018 &\pub{ACCV}  &\makebox[5ex][r]{50}& \makebox[5ex][r]{50}& \makebox[6ex][r]{100} &3.4k & 1 &n/a &\ \ \ \  100&\xmarkg&\xmarkg&\xmarkg\\
\rowcolor[gray]{.96}DAVIS$_{17}$-RVOS~\cite{khoreva2018video} &2018 &\pub{ACCV}  &\makebox[5ex][r]{90}& \makebox[5ex][r]{205}& \makebox[6ex][r]{1,544} &13.5k & 2.27 & 1 &\ \ 1,544& \xmarkg&\xmarkg&\xmarkg\\
Refer-YouTube-VOS~\cite{seo2020urvos} &2020 &\pub{ECCV}  &\makebox[5ex][r]{{3,978}}& \makebox[5ex][r]{7,451}& \makebox[6ex][r]{15,009}& 131k & 1.86 & 1 &15,009&\xmarkg&\xmarkg&\xmarkg\\
\rowcolor[gray]{.96}MeViSv1~\cite{MeViS} &2023 &\pub{ICCV} & \numvideo&  {\numobject}&{28,570} & {443k} & {4.28} & {1.59}&{21,031} &{7,539}&\xmarkg&\xmarkg\\
\hline
\rowcolor[gray]{.9}\textbf{\ourdatasetTPAMI} &2025 & \pub{TPAMI} & \numvideo&  {\numobject}&\textbf{\numsentence} & {443k} & {4.28} & {1.58}&\textbf{{21,541}} &\textbf{{8,028}}&\textbf{{3,503}}&\textbf{{\numsentence}}\\
\hline
\end{tabular}
\end{threeparttable}
\label{table:dataset}
\vspace{-3.6mm}
\end{table*}

\myparagraph{Language Expression Annotation.} 
We developed a web-based annotation system for annotating language expressions in text format. The system randomly selects a video from the \ourdataset dataset and displays all object masks of the selected video on the webpage. For single-target and multi-target expressions, the annotator needs to choose one or several objects from the video and write the corresponding referring expression according to the annotation guidelines. For no-target expressions, the annotator needs to write deceptive expressions without choosing any object. To ensure that the language expressions in \ourdataset align with our focus on motion-based video segmentation, we established several guidelines for annotating the language expressions:
\begin{itemize}
\setlength\itemsep{0em}
\item[A1.] Target objects must exhibit significant motion. Objects that remain stationary and have no motion interactions with other objects should be disregarded.
\item[A2.] If an object can be unambiguously described by its motion or action, static attributes such as shape and color should not be included in the expression.
\item[A3.] If multiple objects cannot be differentiated based solely on their motion or action, they can be described together if their motion or action can unambiguously identify them, such as ``\textit{The two lions fighting and running amidst a group of lions.}"
\item[A4.] If it is not possible to differentiate single or multiple objects based solely on their motion or action, limited static attributes can be included in the expression.
\item[A5.] No-target expression must also describe motion like other single-/multi-target expressions, and cannot be entirely unrelated to the video. Annotators can derive no-target expressions by modifying existing single-/multi-target expressions.
\end{itemize}

\myparagraph{Language Expression Validation.} 
Upon receiving annotated ``video-object-expression'' samples from the annotators, the validation process begins by displaying the video and expression and prompting the validator to select and submit the objects referred to in the expression. The validator must find the targets independently and submit their selection. The system then compares the targets chosen by the validator with the annotations submitted by the annotator. A sample is considered valid if the validator and annotator independently selected the same target object(s) using the same expression. If the targets selected by the validator do not match the annotation submitted by the annotator, the sample will be forwarded to another validator for a second opinion. If the second validator also fails to identify the correct targets, the sample will be considered invalid and excluded from the dataset. Validators have the authority to reject samples that are deemed inappropriate or fall short of quality standards. Moreover, we stress the importance of the following validation criterion:
\begin{itemize}
\item[V1.] The corresponding sentence will be removed from the dataset when the target object described by a sentence can be easily identified through a single frame.
\item[V2.] No-target expressions that are unrelated to the corresponding video or do not describe motion will be discarded.
\end{itemize}
By establishing these validation criteria, we aim to ensure that the language sentences in our dataset accurately express motion and are of high quality, while also increasing the difficulty of the language-guided video segmentation task, thereby enabling more robust evaluation of model performance.

\myparagraph{{Audio Annotation \loudspeaker.}} After the text expressions are created, we further add speech recordings for every language expression. To ensure voice variety, the audio is a mix of automatic synthesis and human recordings. For the human-recorded portion, 10 speakers are employed to read and record the sentences. The speakers come from diverse backgrounds, including native and non-native speakers of different genders and age groups. They are required to read the sentences naturally at a normal talking speed of 100-150 words per minute, and record the audio using microphones with a sampling rate higher than 44.1KHz. Slight pauses, stutters, and background noises are allowed to simulate practical user cases, as long as the speech is recognizable and true to the original text.
For the synthesized portion, we use six state-of-the-art Text-To-Speech (TTS) models and 3 public TTS services. All audio clips are verified twice, by human verifiers and speech recognition models, to ensure they are consistent with the text expression. The total recording time of the dataset is above 150,000 seconds.

\begin{figure}
    \centering
    \includegraphics[width=0.486\textwidth]{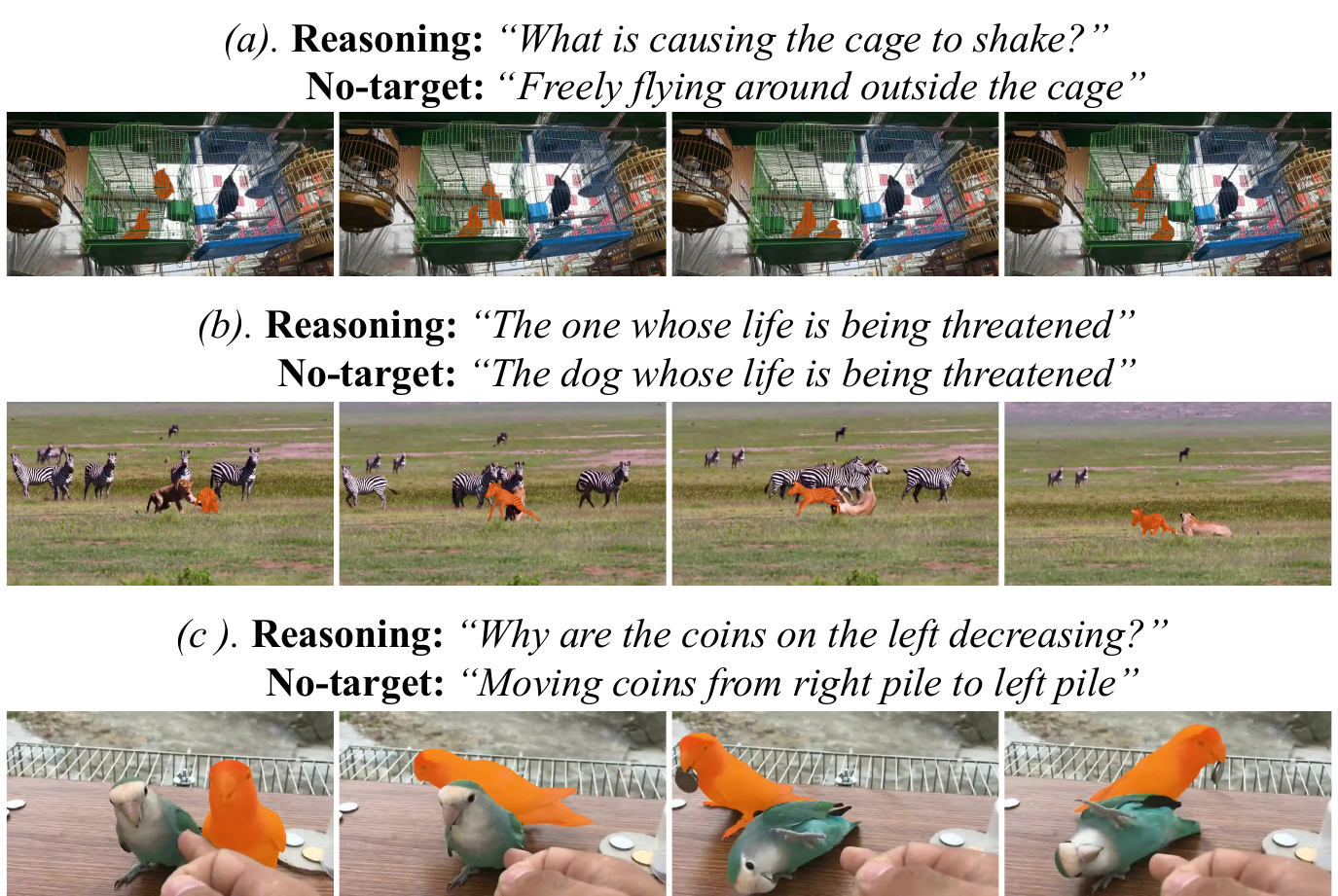}
    \vspace{-5.6mm}
    \caption{Examples of the newly added motion reasoning and no-target expressions in \ourdatasetTPAMI. Motion reasoning expressions refer to the targets the masked in orange, while no-target expressions, though deceptive, do not refer to any objects.}
    \label{fig:motionreasoning}
\end{figure}

\subsection{Dataset Analysis and Statistics}\label{sec:datasetStatistics}

In~\tablename~\ref{table:dataset}, we present a statistical analysis of the newly proposed \textbf{\ourdataset} dataset, using 5 previous referring video object segmentation datasets as references, including A2D Sentence~\cite{gavrilyuk2018actor}, J-HMDB Sentence~\cite{gavrilyuk2018actor}, DAVIS$_{16}$-RVOS~\cite{khoreva2018video}, DAVIS$_{17}$-RVOS~\cite{khoreva2018video}, and Refer-YouTube-VOS~\cite{seo2020urvos}. As shown in \tablename~\ref{table:dataset}, \ourdataset contains \numvideo videos and \numobject objects. Compared to Refer-YouTube-VOS~\cite{seo2020urvos}, which is based on the existing VOS dataset~\cite{xu2018youtube}, \ourdataset has more objects (\numobject~\vs~7,451), more expressions (\numsentence~\vs~15,009), and more annotation masks (443k~\vs~131k). Compared to the previous conference version \ourdatasetICCV~\cite{MeViS}, \ourdatasetTPAMI offers more expressions (\numsentence~\vs~28,570) by including: 1) 999 motion reasoning expressions that implicitly refer to the target through complex motion reasoning, and 2) 3,503 no-target expressions that are deceptive but do not refer to any objects in the video, as shown in \cref{fig:motionreasoning}. Motion reasoning expression necessitates reasoning based on implicit motion clues, while no-target expressions support the robustness study in language-guided video segmentation. Both additions expand the real-world applications of the MeViS dataset. Moreover, \ourdatasetTPAMI provides audio format referring expressions to support multi-modal studies in the field. In the following, we discuss how the proposed dataset \ourdataset intentionally increases the complexities of language-guided video segmentation by considering the challenges of both linguistic and visual modalities.

\begin{figure}
    \centering
    \hspace{-5mm}
    \includegraphics[width=0.25\textwidth]{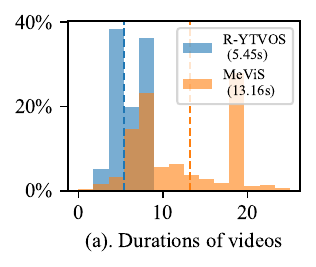}
    \hspace{-3mm}
    \includegraphics[width=0.25\textwidth]{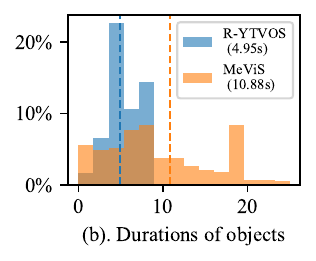}
    \hspace{-6mm}
    \vspace{-3mm}
    \caption{The duration of videos and objects of \ourdataset and Refer-YouTube-VOS~\cite{seo2020urvos}, in seconds. The vertical lines and values in the legends represent the mean duration across the two datasets. The duration of both videos and objects in \ourdataset is significantly longer than Refer-YouTube-VOS.}
    \label{fig:VideoContent2}
    \vspace{-2mm}
\end{figure}

\begin{figure}
    \centering
    \includegraphics[width=0.476\textwidth]{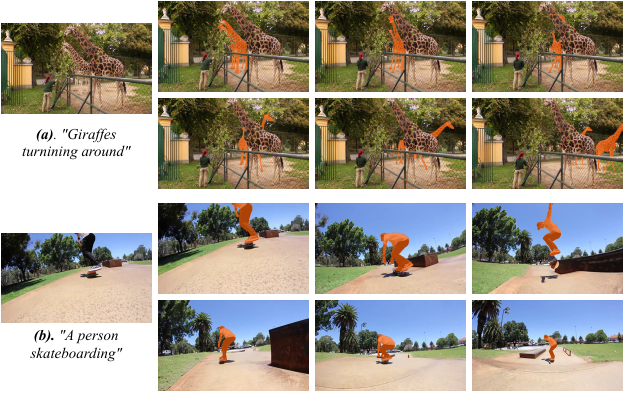}
    \vspace{-1.6mm}
    \caption{Comparison of MeViS and Refer-YouTube-VOS~\cite{seo2020urvos}: (a) Example from \ourdataset. (b) Example from Refer-YouTube-VOS~\cite{seo2020urvos}. Compared to Refer-YouTube-VOS: $\bullet$ Videos in \ourdataset contain \textbf{more objects} in complex environments, making it impossible to identify the target object via saliency or category information alone. $\bullet$ The number of target objects indicated by language expression in \ourdataset is \textbf{arbitrary, from 0 to many}.}
    \label{fig:VideoContent}
\end{figure}

\textbf{More Challenging Videos.} 
As shown in \tablename~\ref{table:dataset}, \ourdataset has an average of 4.28 objects per video. This is significantly higher than all previous datasets and is more than twice the number in the largest previous dataset, Refer-YouTube-VOS. The increased number of objects per video introduces more complex relationships among objects and poses greater challenges for understanding video content. 
Furthermore, as shown in \figurename~\ref{fig:VideoContent2}, \ourdataset contains considerably longer videos, with an average duration of 13.16 seconds, which is significantly longer than the 5.45 seconds of Refer-YouTube-VOS~\cite{seo2020urvos} dataset. These long-term videos introduce unique challenges, such as frequent disappearance-reappearance and prolonged confusion among similar-looking objects. 
These intentional design choices make \ourdataset more complex and challenging for language-guided video segmentation. This is in contrast to existing datasets such as A2D Sentence~\cite{gavrilyuk2018actor} and DAVIS$_{16}$-RVOS~\cite{khoreva2018video}, where only one or two salient objects per category are present, and the model can choose the most salient object as the target or identify the target object based on the category name. 
For example, in \figurename~\ref{fig:VideoContent}(b), there is only one person in the foreground, and the model can simply identify the target by the term {``\textit{a person}''} while ignoring {``\textit{skateboarding}''}. 
The proposed \ourdataset dataset addresses this limitation by selecting videos with more objects that have diverse and dynamic motions. 
Moreover, \ourdataset includes many videos with objects of the same category, such as a group of tigers or rabbits. By including more challenging videos, \ourdataset better simulates real-world scenarios, making it a valuable resource for studying motion expression-guided video understanding in complex environments.

\textbf{More Challenging Target Objects.} 
As we have included longer videos in our \ourdataset dataset, we have also observed a significant increase in the duration of target objects. As shown in \figurename~\ref{fig:VideoContent2}(b), the object durations in our dataset have an average of 10.88 seconds, which is more than two times longer than the average duration of Refer-YouTube-VOS. The longer duration of target objects ensures sufficient object motions and increases the difficulty of motion understanding. 
In previous datasets, target objects are typically salient, dominant, and isolated. For example, in \figurename~\ref{fig:VideoContent}(b), the target person is the absolute only protagonist of the video. Inconspicuous objects are rarely the referred targets, which does not align with real-world applications. In contrast, the proposed MeViS dataset includes numerous targets that are inconspicuous, small, entangled with other objects, or in the background.
For example, in \figurename~\ref{fig:VideoContent}(a), there are three giraffes with highly similar appearances, and the most salient/foreground one is not the target object in this sample, making it challenging to identify the target object(s) through saliency or category information alone.
Then compared to previous datasets, such as A2D Sentence~\cite{gavrilyuk2018actor} and J-HMDB Sentence~\cite{gavrilyuk2018actor}, which focus on a few categories~\cite{seo2020urvos}, our \ourdataset dataset includes more categories from open-world~\cite{UVO,OVIS,MOSE,TAOVOS,MOSEv2}, presenting improved difficulties in the diversity of target objects. 

\textbf{Generalized Referring Expressions.} As shown in the ``Target Object/Expression" of \tablename~\ref{table:dataset}, previous datasets typically have one sentence referring to a single object, \ie, ``{\textit{one expression, one object}}'' has become a ``de-facto'' rule. This implies that finding multiple objects requires multiple expressions, with each object being searched for individually.
In contrast, we add a more natural way of selecting target objects, allowing one expression to refer to multiple target objects, denoted as ``multi-target expression''. An example of multi-target expression is shown in \figurename~\ref{fig:VideoContent}(a), where \textit{``Giraffes turning around''} refers to two giraffes. As shown in \tablename~\ref{table:dataset}, on average, each expression in \ourdatasetICCV refers to 1.59 objects, which is larger than existing datasets where the average is only 1 object per expression. However, in previous version ``no-target expression'' is not considered, leading to undefined behavior when the given sentence does not match any object in the given video. To address this issue and enhance the practical applications of referring video segmentation, we further add no-target expressions through human annotation in \ourdatasetTPAMI, especially focusing on motion confusion like \textit{``Moving coins from right pile to left pile''} in \cref{fig:motionreasoning}. Allowing multi-target and no-target expressions makes \ourdataset more practical and generalized to real-world scenarios. Generalized referring expressions~\cite{GRES} help to enhance the model’s reliability and robustness in realistic scenarios, where any type of expression can occur unexpectedly.

\begin{figure}
    \centering
    \includegraphics[width=0.46\textwidth]{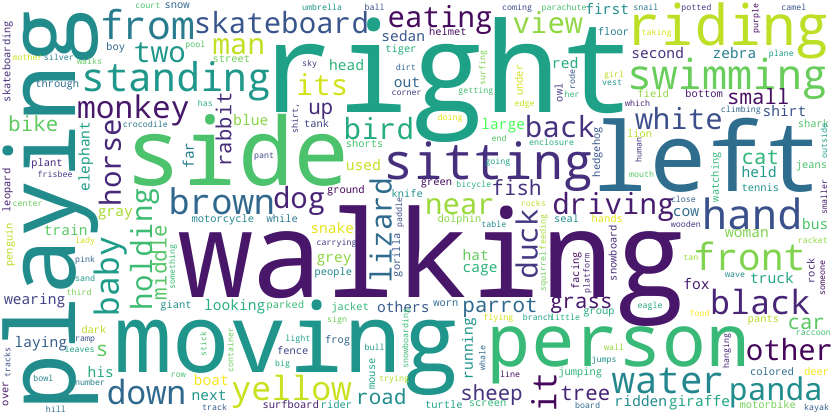}
    \caption{Word cloud of the top 100 words in the \ourdataset dataset. \ourdataset has a large number of words that describe motions, like ``\textit{walking}'',  ``\textit{moving}'', ``\textit{playing}'', and many position words that are related to motions, such as ``\textit{left}'', ``\textit{right}''.}
    \label{fig:wordcloud}
\end{figure}

\textbf{More Challenging Motion Clues.} 
One of the key distinguishing aspects of the \ourdataset dataset is its emphasis on describing object motions in language expressions. The previous largest RVOS dataset Refer-YouTube-VOS~\cite{seo2020urvos} provides two types of language annotations: full-video expression and first-frame expression. The first-frame expression is based solely on static attributes of the first frame image, whereas the full-video expression considers the entire video. However, in many cases, even the full-video expressions contain static attributes that could potentially enable the target object to be identified in a single frame, for example, \textit{``A person on the right dressed in {blue black}...''}. In contrast, to explore the practicality of employing motion expressions for object localization and segmentation in videos, \ourdataset is intentionally designed to include a range of diverse and dynamic object motions, making it more challenging to identify the target object based on static attributes alone. In \ourdataset, there are significantly more motion expressions that explicitly identify the target object based on its distinctive actions or movements. The language expressions in the proposed \ourdataset contain more motion attributes, such as object position moving through the video and actions that span several frames. The word cloud of the newly proposed \ourdataset is visualized in \figurename~\ref{fig:wordcloud}. From the word cloud figure, we can observe that \ourdataset dataset has a large number of words that describe motions, like \textit{``walking''}, \textit{``moving''}, \textit{``playing''}, and many relative directions that are related to motions, such as \textit{``left''}, \textit{``right''}, etc.

\textbf{Multi-modal Referring Expressions: Text~\&~Audio.} Besides the text-based referring expressions, we further add audio-based referring expressions in the updated \ourdatasetTPAMI.  
Audio, as a reflection of human cognition, is more natural, common, and convenient in daily interactions compared to text. It carries rich semantic information and captures nuances of tone, emotion, and emphasis that text alone cannot convey. These qualities aid in more precise target identification and segmentation. The newly added audio format in \ourdatasetTPAMI supports not only audio-guided video object segmentation but also multi-modal referring expression tasks.
By leveraging the strengths of both text and audio, multi-modal referring expressions offer significant advantages and flexibility in enhancing video understanding and supporting more natural and intuitive interactions.

\subsection{Tasks Supported by MeViS}\label{sec:supportedTasks}
\textbf{1) Referring Video Object Segmentation (RVOS).} The proposed MeViS dataset is originally designed for referring video object segmentation (RVOS), emphasizing motion understanding of both linguistic and visual contents. Besides RVOS, MeViS is versatile and applicable to a variety of other tasks, as outlined below.

\textbf{2) Audio-guided Video Object Segmentation (AVOS).}
As described in \cref{sec:videoannotation}, we further provide corresponding speech recordings to each textual expression in \ourdatasetTPAMI, enabling MeViS to be used for Audio-guided Video Object Segmentation (AVOS) \cite{pan2022wnet, lin2024echotrack}. This task suits future embodied scenarios, where using speech to command a robot is more convenient than inputting text. An intuitive solution is directly using automatic speech recognition models to convert audio to text, thus degenerating it to the aforementioned RVOS. However, this method overlooks the rich semantic information in audio, such as accent, emotion, speed, and noise \cite{pan2022wnet}. Recent works \cite{zhan2024anygpt, pan2022wnet} demonstrate the potential of using audio as a single modality without the need for text as an intermediary. 
Therefore, directly integrating audio with visual signals for achieving efficient referring segmentation is a good direction.
Additionally, compared to earlier AVOS datasets \cite{lin2024echotrack, pan2022wnet}, which are extended from simple RVOS \cite{jhuang2013towards, A2D, seo2020urvos} and have the drawback of relying on target saliency that could be judged from a single frame, the audio version of MeViS inherits the complex relationships between expressions and targets from MeViS. This increases the task's complexity and evaluates the model's generalization in real-world scenarios.

\begin{figure*}[t]
    \centering
    \includegraphics[width=0.999\textwidth]{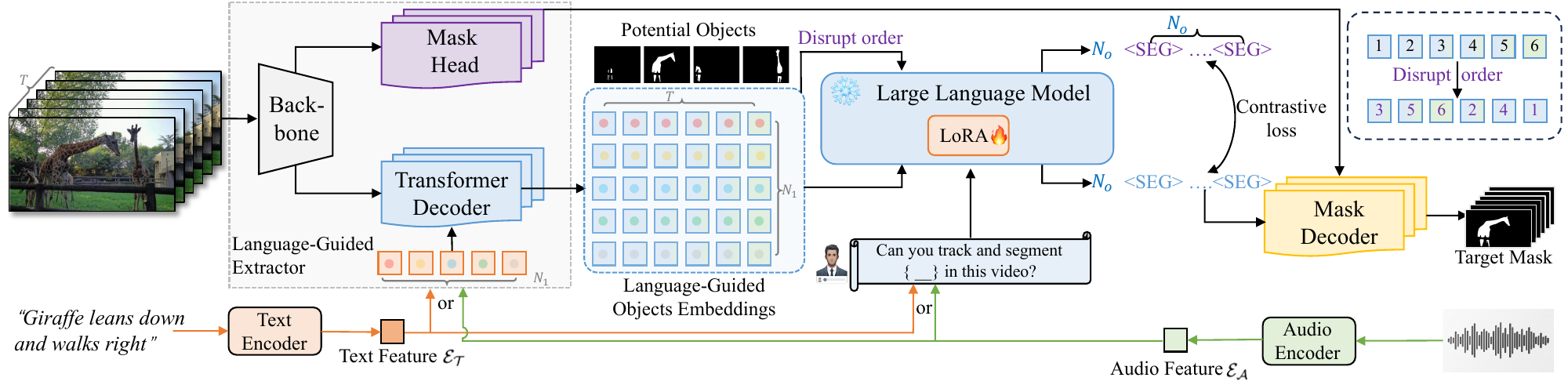}
    \vspace{-5.6mm}
    \caption{The overview architecture of the proposed Language-guided Motion Perception and Matching (\textbf{\ournewmodel}). We first detect all possible target objects in each video frame via Language-Guided Extractor and represent them using object embeddings. Then, a large language model is used to capture and reason the global temporal context from object embeddings, outputting the number of target objects $N_o$ and the corresponding $N_o$ \texttt{<SEG>} tokens for subsequent mask generation. Temporal-level contrastive loss is designed to enhance the understanding of temporal structure.
    Finally, object mask trajectories are generated using \texttt{<SEG>} token with Mask Decoder.}
    \label{fig:L-MAM}
\end{figure*}

\textbf{3) Referring Multi-Object Tracking (RMOT).}
RMOT aims to detect and track objects by generating bounding box trajectories based on natural language descriptions. MeViS can be seamlessly adapted to RMOT by converting segmentation masks into bounding boxes. Unlike previous RMOT datasets such as Refer-KITTI \cite{rmot}, which focus on autonomous driving, MeViS includes a wider variety of scenes, enhancing its relevance to real-world applications and its ability to test model generalization. Additionally, MeViS introduces several unique features, such as the no-target expressions and an emphasis on understanding long-term actions through natural language expressions. Moreover, MeViS offers a significantly larger dataset, comprising 2,006 videos, 136,102 frames, and \numsentence expressions, in comparison to Refer-KITTI’s \cite{rmot} 18 videos, 6,650 frames, and 818 expressions. This makes MeViS better suited for large-scale model training.

\textbf{4) Referring Motion Expression Generation (RMEG)}. Besides the aforementioned perception tasks, the MeViS dataset is also suitable for language generation tasks. 
One of the relevant tasks is video captioning~\cite{fu2023empirical,yamazaki2022vlcap,wang2022git}, which aims to generate descriptive expressions given a video. Traditional video captioning can be divided into two categories: single sentence video captioning~\cite{gao2017video,pei2019memory} and dense video captioning~\cite{suin2020efficient,xu2023mplug,xu2019joint,zhou2023dense,yang2023vid2seq}.
The former one requires to generate one expression that describes the video globally, which is only applicable for videos with an explicit theme or a salient subject. In contrast, dense video captioning methods generate multiple expressions that caption multiple events~\cite{xu2019joint} or objects\cite{zhou2023dense} in the video.
However, they both focus on ``description'' without the need to ``differentiate'' between different objects or events, meaning that the generated expressions are not unambiguously affiliated with the object.

Different from traditional video captioning, we propose a new task with the help of the MeViS dataset, namely \textbf{Referring Motion Expression Generation} (RMEG). 
The input of RMEG is a video along with a set of masks of specific target object(s) in this video. The model is expected to generate a referring expression that unambiguously describes the target's motion and distinguishes it from other objects. This poses a higher challenge to the methods regarding their scene and object understanding capability.
Recently, object-oriented~\cite{liu2021o2na} and controllable~\cite{cornia2019show} video captioning are proposed. However, they require an explicit object list input to decide which objects will appear in the output expression. In RMEG, we expect the model to find all relevant objects and form an appropriate expression by itself.

\textbf{5) Applications in Additional Tasks such as AIGC and Beyond.} 
\ourdatasetTPAMI includes referring expressions in both text and audio formats, along with segmentation masks and bounding boxes, making it applicable to a wide range of areas. 
We have already seen MeViS being used for tasks beyond those previously mentioned. For example, VIDiff~\cite{xing2023vidiff} employs our dataset to train diffusion-based models for generative video editing, enabling the modification and translation of content based on user instructions. Merlin~\cite{yuen2023merlin} uses our dataset to train Multi-modal Large Language Models (MLLMs) that can foresee the future based on present observations. These examples demonstrate the extensive potential and versatility of our dataset in various applications.
\section{\ournewmodel: A Baseline Approach}
\label{sec:L-TAM}

The \ourdataset dataset introduces unique challenges in detecting and understanding object motions in both video and language contexts. Motions described by language expressions can occur over a random number of frames, making it necessary to capture fleeting actions and movements that occur throughout the entire video. This presents significant challenges for recognizing motions in the video content and the corresponding language expressions. Detecting fleeting actions requires meticulous perceiving of every frame while comprehending complex and extended motion spanning multiple frames requires contextual understanding across the entire duration of the video. Current state-of-the-art methods~\cite{wu2022referformer,MTTR,Ding_2022_CVPR} rely on random sampling of a few frames, which may miss frames containing crucial information described by the given expression. Furthermore, these methods fail to effectively extract temporal contextual information and instead simply use spatial-temporal feature extractors due to the significant burden on computational resources of temporal communication. Additionally, as illustrated in \cref{sec:MeViS_Dataset}, objects described by language expressions can vary from zero to multiple, requiring the output to cover from zero to an arbitrary number of objects.

To address the challenges posed by \ourdataset, we propose a new approach called Language-guided Motion Perception and Matching (\ournewmodel), as shown in~\cref{fig:L-MAM}. \ournewmodel generates $N_1$ language-based queries to identify potential target objects in the video, across $T$ frames, and produces object embeddings to represent each of them. Using language queries instead of conventional object queries can filter out irrelevant objects and ensure the efficiency and effectiveness of subsequent operations~\cite{ding2021vision,vltpami}. 
Inspired by VITA~\cite{VITA}, we represent objects using object embeddings, which provide instance-specific information, to reduce computational requirements~\cite{li2023transformer, li2023tube}. 
Recent developed large language models (LLM) have the ability to reason about complex sentences~\cite{llama2,vicuna2023,LLaVA,LISA}. Given their strengths, we employ LLM for motion modeling of both long-term motion and fleeting motion.
After obtaining object embeddings from frames in the video, 
we utilize LLM $\mathcal{E}$~\cite{llama2,vicuna2023} to obtain a global view across T frames with LoRA~\cite{LoRA} fine-tuning strategy.

To support multimodal input, we design separate feature extraction branches for text and audio.
For text prompts, we introduce \texttt{<Text>} and \texttt{</Text>} for LLM to recognize text-referring inputs.
Extracted text features are projected through a text-language projection layer and inserted between prompt tokens, forming the instruction: “Can you track and segment \texttt{<Text><Text Embedding></Text>} in this video?” The text embedding is generated by the Text Encoder $\mathcal{E_T}$.
Similarly, for audio prompts, we use \texttt{<Audio>} and \texttt{</Audio>} tags, with audio features processed through an audio-language projection layer and inserted into prompt tokens. The instruction is: “Can you track and segment \texttt{<Audio><Audio Embedding></Audio>} in this video?” Audio embedding is produced by the Audio Encoder $\mathcal{E_A}$.
In this way it unifies referring representations across modalities, enabling the LLM to handle them like language instructions. 

To deal with the newly added generalized referring expressions in \ourdatasetTPAMI and further support multi-target and no-target outputs, we design the answer template of LLM as: ``{$N_o$} \texttt{<SEG>$\cdots$}''. $N_o$ denotes the predicted number of target objects as well as the number of \texttt{<SEG>} tokens. If $N_o$ = 0, there is no target object, and $N_o>1$ indicates multiple target objects. This answer template facilitates the LLM to better understand the generalized referring expressions and enhances the practicality of \ournewmodel.

Another challenge is how to facilitate LLM to understand visual-temporal information, such as ``\textit{first jumping high and then jumping far}'' or ``\textit{first jumping far and then jumping high}''. The model struggles to effectively construct temporal structure from the input object embeddings, leading to potential ambiguities and false positives during inference. To address this issue, we propose a temporal-level contrastive loss to encode temporal knowledge. First, we randomly disrupt the object embeddings along the time axis, breaking the original sequential order. We then apply contrastive learning to distinguish the target \texttt{<SEG>} token from the disrupted \texttt{<SEG>} token. This is done by maximizing the similarity between the target \texttt{<SEG>} and the corresponding text or audio expression while minimizing the similarity between the disrupted token and its associated text or audio expression. Additionally, to increase the number of samples, we gather tokens across different GPUs and incorporate them into the process.
The process can be formulated as multi-positive supervised contrastive learning:
\begin{equation}\label{eqn:inter}
    \mathcal{L}_{con} = -\frac{1}{|\mathbf{P}|}\sum_{a^+\in \mathbf{P}}\log\frac{\mathrm{exp}(<{a},{a^+}>/\tau)}{\sum_{a'\in \mathbf{P},\mathbf{N}}\mathrm{exp}(<{a},{a'}>/\tau)},
\end{equation}
where $a$ represents the anchor item, a text or audio embedding, and $a^+$ and $a'$ are \texttt{<SEG>} embeddings. 
$\mathbf{P}$ is the collection of positive samples matched to the language embedding. 
$\mathbf{N}$ is the collection of negative samples that come from different objects and disrupted embeddings. 
The primary goal of $\mathcal{L}_{con}$ is to refine the embedding space so that text or audio embedding align with the \texttt{<SEG>} embeddings that have the correct temporal order, while staying distanced from embeddings with incorrect temporal orders or embeddings of different objects. This approach effectively mitigates ambiguities caused by distractor embeddings and temporal inconsistencies, and significantly enhances the referring video understanding performance.

Our model is trained end-to-end using a combination of text classification loss, contrastive loss, and segmentation mask loss:
\begin{equation}
    \mathcal{L} = \lambda_{txt} \mathcal{L}_{txt} + \lambda_{bce} \mathcal{L}_{bce}+ \lambda_{dice} \mathcal{L}_{dice} + \lambda_{con}\mathcal{L}_{con} ,
\end{equation}
{where $\mathcal{L}_{txt}$ denotes the auto-regressive cross-entropy loss, optimizing text generation accuracy for $N_o$ and \texttt{<SEG>} tokens. The segmentation loss includes binary cross-entropy loss $\mathcal{L}_{bce}$ and DICE loss $\mathcal{L}_{dice}$, which work together to enhance segmentation quality. These loss items are balanced by the weights $\lambda_{txt}$, $\lambda_{bce}$, $\lambda_{dice}$, and $\lambda_{con}$. During training, the model is guided by the ground-truth labels $\vec{y}_{txt}$ for text and $\vec{M}$ for segmentation masks.}
\section{Experiments}

\begin{table}[t]
    \centering
     \footnotesize
     \setlength\tabcolsep{7pt}
       \caption{Temporal Context ({TC}) shows varying impacts on 3 datasets. Image-based methods, like VLT~\cite{vltpami}, can achieve state-of-the-art performance on DAVIS$_{17}$-RVOS (D$_{17}$R)~\cite{khoreva2018video} and Refer-YouTube-VOS (RYV)~\cite{seo2020urvos}, but cannot well handle the harder motion challenges in \ourdataset that require temporal context.}\label{tab:MeViSnecessarity}
  \vspace{-2mm}
   \setlength{\tabcolsep}{2.36mm}{\begin{tabular}{lccccc}
         \specialrule{.1em}{.05em}{.05em}
         \rowcolor[gray]{.9}Methods& Type&Temporal& D$_{17}$R&RYV & MeViS\\\hline\hline
         VLT~\cite{vltpami}&\textbf{Image}&1 frame&60.4&63.1&27.8\\
         RFormer~\cite{wu2022referformer} &Video&5 rand. frames&60.2&62.8&31.0\\
         VLT+TC&Video&All frames&60.3&62.7&35.5\\
         RFormer+TC&Video&All frames&59.9&63.0&36.3\\
         \specialrule{.1em}{.05em}{.05em}
      \end{tabular}}%
      \vspace{-0.6mm}
\end{table}

\begin{table}[t]
    \footnotesize
   \centering
    \caption{Image-video cross-dataset validation. We train the models on referring image segmentation dataset Ref-COCO/+/g, and test their performance on video datasets. The models trained on images perform worse on \ourdataset than on the other two datasets.}\label{tab:MeViSnecessarity2}
  \vspace{-2mm}
  \renewcommand\arraystretch{1.16}
   \setlength{\tabcolsep}{3.16mm}{\begin{tabular}{lcccc}
         \specialrule{.1em}{.05em}{.05em}
         \rowcolor[gray]{.9}\multicolumn{5}{c}{Training on Referring Image Segmentation Dataset}\\\hline
         Methods& Type& DAVIS$_{17}$-RVOS&RYV & MeViS\\
         \hline 
         VLT~\cite{vltpami}&Image& 54.2 & 46.1 & 22.5 \\
         RFormer~\cite{wu2022referformer} &Video&55.6&45.2&27.0\\
         \specialrule{.1em}{.05em}{.05em}
      \end{tabular}}%
    \vspace{-1.6mm}
\end{table}

\myparagraph{Dataset Setting.} The proposed \ourdatasetTPAMI dataset consists of a total of \numvideo videos along with \numsentence sentences. These videos are split into three subsets, \ie, training set, validation set for daily online evaluation, and testing set for competition~\!\footnote{The testing set is used for evaluation during the competition periods, such as \url{https://lsvos.github.io/} and \url{https://pvuw.github.io/}.}, which contain 1,712 videos, 140 videos, and 154 videos, respectively.

\myparagraph{Evaluation Metrics.} For RVOS and AVOS tasks, following~\cite{khoreva2018video,seo2020urvos}, we employ \(\mathcal{J}\) and \(\mathcal{F}\) to assess segmentation performance on the newly proposed \ourdataset dataset. The region similarity metric \(\mathcal{J}\) computes the Intersection over Union (IoU) of the predicted and ground-truth masks, reflecting the segmentation quality. The F-measure \(\mathcal{F}\) reflects the contour accuracy of the prediction. To evaluate the overall performance, we calculate the average of these two metrics, denoted as \(\mathcal{J\&F}\).
It is worth noting that for samples where no target is present, true positives are assigned \(\mathcal{J}\) and \(\mathcal{F}\) value of 1, whereas false negatives are given \(\mathcal{J}\) and \(\mathcal{F}\) value of 0. Besides, {N-acc. and T-acc.}~\cite{GRES} are employed to assess the model's ability to identify no-target scenarios. 
N-acc. (No-target accuracy) measures the model’s ability to identify no-target samples: $\text{N-acc.}$~\!=~\!$\frac{\mathit{TP}}{\mathit{TP} + \mathit{FN}}$, where $\mathit{TP}$ is the number of correctly identified no-target samples and $\mathit{FN}$ is the number of no-target samples misclassified as target samples. T-acc. (Target accuracy) reflects how no-target generalization affects target performance: $\text{T-acc.}$~\!=~\!$\frac{\mathit{TN}}{\mathit{TN} + \mathit{FP}}$, where $\mathit{TN}$ is the number of correctly identified target samples and $\mathit{FP}$ is the number of target samples misclassified as no-target.
For the evaluation metrics of RMOT and RMEG tasks, please refer to Sec.~\ref{sec:eval_RMOT} and Sec.~\ref{sec:eval_RMEG}, respectively.

\myparagraph{Implementation Details.} We set all the hyper-parameters of Language-Guided Extractor including Backbone, Mask Head, and Transformer Decoder to the default settings of Mask2Former \cite{mask2former}. We train 150,000 iterations using AdamW optimizer \cite{loshchilov2017adamw} with a learning rate of 0.00005.
Tiny Swin Transformer~\cite{liu2021swin} is used as our backbone. The input frames are resized to have a minimum size of 360 pixels on the shorter side and a maximum size of 640 pixels on the longer side, to ensure efficient memory usage on the GPU. We use Video-LLaMA-7B \cite{videollama} as our large language model,  with its vision encoder removed. For audio encoding, we employ the pre-trained ImageBind \cite{imagebind}, following Video-LLaMA’s processing pipeline. Our model is trained end-to-end on 8 NVIDIA A6000 GPUs.
For the hyper-parameter settings, we set $N_1$, $\lambda_{txt}$, $\lambda_{bce}$, $\lambda_{dice}$, and $\lambda_{con}$ to 20, 1, 2, 0.5, and 0.3, respectively. We use RoBERTa~\cite{liu2019roberta} as a text encoder that is consistent with the ReferFormer and is frozen all the time.

\begin{table}[t]
   \renewcommand\arraystretch{1.16}
   \centering
   \footnotesize
   \caption{Ablation study of \ournewmodel in $\mathcal{J\&F}$.}
  \vspace{-2mm}
   \setlength{\tabcolsep}{1.36mm}{\begin{tabular}{cccccc}
         \specialrule{.1em}{.05em}{.05em}
        \rowcolor[gray]{.9} ID&LMPM&Large language model & $\mathcal{L}_{con}$ & \ourdatasetICCV & \ourdatasetTPAMI \\
         \hline\hline 
         \romannumeral1&\cmark&\xmarkg&\xmarkg&42.2 &38.3\\
         \romannumeral2&\cmark&\cmark&\xmarkg&45.9 &42.4\\
         \romannumeral3&\cmark&\cmark&\cmark&47.6 &43.9\\
         \specialrule{.1em}{.05em}{.05em}
      \end{tabular}}%
   \label{tab:TAM}%
\end{table}%

\begin{table}[t]
\centering
\footnotesize
\setlength{\tabcolsep}{0.6mm}
\caption{Computational cost comparison on \ourdatasetTPAMI.}
\vspace{-2mm}
{\begin{tabular}{l|ccc|ccc}
         \specialrule{.1em}{.05em}{.05em}
        \rowcolor[gray]{.9} Methods& Backbone & FPS& \#Params.& $\mathcal{J}$\&$\mathcal{F}$ & N-Acc.&T-Acc \\
         \hline
         \hline
         LMPM~\cite{MeViS} & Swin-T &32.1 & 66.39M&{38.3}&20.1 & 72.9 \\ 
         VISA~\cite{VISA}& Chat-UniVi-13B & 1.72&22.98B&40.7&\ \ 1.9&87.2\\
         \hline
         \textbf{LMPM++} (ours)& Video-LLaMA-7B &2.86&\ \ 7.08B &\textbf{43.9}&\textbf{45.7}& \textbf{87.4} \\
         \specialrule{.1em}{.05em}{.05em}
      \end{tabular}}
      \vspace{-1.6mm}
   \label{tab:MeViS_efficiency}
\end{table}

\subsection{Dataset Necessity and Challenges}

To show the necessity and validity of \ourdataset in motion expression understanding, we compare the results of state-of-the-art referring image segmentation method VLT~\cite{vltpami} and referring video segmentation method ReferFormer~\cite{wu2022referformer} on DAVIS$_{17}$-RVOS~\cite{khoreva2018video}, Refer-YouTube-VOS~\cite{seo2020urvos}, and \ourdataset, as shown in \cref{tab:MeViSnecessarity}. When trained on referring video segmentation dataset, such as Refer-YouTube-VOS~\cite{seo2020urvos} and testing on itself, the image-based method VLT~\cite{vltpami} that does not use any temporal design can achieve exceptional results of 60.4\% $\mathcal{J\&F}$ and 63.1\% $\mathcal{J\&F}$ on video datasets DAVIS$_{17}$-RVOS~\cite{khoreva2018video} and Refer-YouTube-VOS~\cite{seo2020urvos}, respectively, which are even better than video method ReferFormer~\cite{wu2022referformer}. The results suggest that for DAVIS$_{17}$-RVOS~\cite{khoreva2018video} and Refer-YouTube-VOS~\cite{seo2020urvos}, the temporal context is not essential, and image-based methods that use static clues can achieve good performance on these two datasets. However, on the proposed \ourdataset, VLT~\cite{vltpami} only achieves a score of 27.8\% $\mathcal{J\&F}$, suggesting that referring image segmentation methods without temporal designs struggle to address the unique challenges presented by videos in our dataset, particularly in handling motion, despite their success on other benchmark datasets. Furthermore, by comparing the results of VLT~\cite{vltpami} with ReferFormer~\cite{wu2022referformer}, which is trained using five randomly selected frames from the video, we find that ReferFormer outperforms VLT by a large margin of {3.2\%} in terms of $\mathcal{J\&F}$. This further highlights the importance of analyzing long-term motions in the \ourdataset dataset. In order to further prove this point, we enhance VLT and ReferFormer by incorporating an attention module at the head to perceive and gather global temporal context (``TC'' in \cref{tab:MeViSnecessarity}). 
Adding temporal context results in both VLT and ReferFormer achieving a performance gain of approximately 5\% $\mathcal{J\&F}$, underscoring the significance of temporal context for \ourdataset. However, it is worth noting that longer temporal information does not necessarily lead to better performance on DAVIS$_{17}$-RVOS and Refer-YouTube-VOS.

We also conduct a cross-dataset experiment, where models are trained on referring image datasets and tested on referring video datasets. As shown in \cref{tab:MeViSnecessarity2}, both the image-based method VLT~\cite{vltpami} and video-based method ReferFormer~\cite{wu2022referformer} achieve competitive results on Refer-YouTube-VOS~\cite{seo2020urvos} and DAVIS$_{17}$-RVOS~\cite{khoreva2018video} when trained on image datasets Ref-COCO, Ref-COCO+, and Ref-COCOg. These results suggest that the expressions in Refer-YouTube-VOS~\cite{seo2020urvos} and DAVIS$_{17}$-RVOS~\cite{khoreva2018video} provide static clues like in the image domain, and many target objects can be identified by examining a single frame solely. In contrast, when trained on referring image segmentation datasets and tested on \ourdataset, both VLT~\cite{vltpami} and ReferFormer~\cite{wu2022referformer} perform worse, indicating that there is a significant expression-gap (\eg, static \vs motion) between \ourdataset and these image domain datasets.

\subsection{Ablation Study of \ournewmodel} 
In \cref{tab:TAM}, we present an ablation study of the proposed approach \ournewmodel. We conduct the following three experiments:
(\romannumeral1) First, we utilize language queries as cues to detect potential target object trajectories and capture temporal context through motion aggregation, outputting trajectories whose similarity to language is greater than a matching threshold. This variant, denoted as LMPM~\cite{MeViS}, achieves a $\mathcal{J\&F}$ score of 42.2\% on \ourdatasetICCV and 38.3\% on \ourdatasetTPAMI. 
However, LMPM struggles to understand implicit language information in motion reasoning expressions. Additionally, despite using a cross-attention mechanism to gather temporal context, it overlooks the temporal sequential order within the video. These limitations reduce its effectiveness in processing long-term and complex motions.
(\romannumeral2) By incorporating a large language model for motion perception, which includes predicting the number of objects as part of the output, the $\mathcal{J\&F}$ score improves significantly by 3.7\% and 4.1\% on \ourdatasetICCV and \ourdatasetTPAMI, respectively. This improvement is due to the model's ability to capture temporal contextual information and embed reasoning capabilities, which are critical for \ourdataset. Additionally, this variant enables the model to handle single-object, no-target, and multi-object expressions effectively.
(\romannumeral3) Given that \ourdataset contains motion expressions within video sequences, it is crucial for the model to understand temporal motion sequence. To enhance this capability, we introduce a temporal-level contrastive loss $\mathcal{L}_{con}$, which further improves the model's comprehension of temporal information. This variant outperforms (\romannumeral2) by 1.7\% and 1.5\% $\mathcal{J\&F}$ on \ourdatasetICCV and \ourdatasetTPAMI, respectively.

\tablename~\ref{tab:MeViS_efficiency} compares computational costs. LMPM++ introduces only a slight increase in trainable parameters (+12.76M) over LMPM~\cite{MeViS} due to the use of LoRA, though with slower inference speed. Compared to LLM-based methods such as VISA~\cite{VISA}, LMPM++ is both faster and more parameter-efficient, as it leverages object tokens rather than frame-level features for handling long video sequences.

\begin{table}[t]
   \renewcommand\arraystretch{1.16}
   \centering
   \footnotesize
   \caption{\ourdatasetICCV Benchmark Results.}
  \vspace{-2mm}
   \setlength{\tabcolsep}{3.76mm}{\begin{tabular}{l|c|ccc}
         \specialrule{.1em}{.05em}{.05em}
          \rowcolor[gray]{.9}
\rowcolor[gray]{.9} 
        \rowcolor[gray]{.9} Methods& Reference&$\mathcal{J\&F}$ & $\mathcal{J}$ & $\mathcal{F}$\\
         \hline\hline
         URVOS~\cite{seo2020urvos}& \pub{ECCV'20} &27.8&25.7&29.9\\
         LBDT~\cite{Ding_2022_CVPR}&\pub{CVPR'22}&29.3&27.8&30.8\\
         MTTR~\cite{MTTR}&\pub{CVPR'22}&30.0&28.8&31.2\\
         ReferFormer~\cite{wu2022referformer}&\pub{CVPR'22} &31.0&29.8&32.2\\
         VLT+TC~\cite{vltpami}&\pub{TPAMI'23}&35.5&33.6&37.3\\
         DsHmp~\cite{DsHmp}&\pub{CVPR'24}&46.4& 43.0 &49.8 \\
         \hline         
         \textbf{\ourmodel} (ours) &\pub{ICCV'23}& {42.2} &{39.3}& {45.1}\\
          \textbf{\ournewmodel} (ours) &\pub{TPAMI}& \textbf{47.6} &\textbf{44.3}& \textbf{50.9}\\
         \specialrule{.1em}{.05em}{.05em}
      \end{tabular}}%
   \label{tab:MeViS2023}%
\end{table}

\subsection{\ourdataset Benchmark Results}

\myparagraph{Quantitative Results.}~We conduct a comprehensive evaluation of the \ourdataset dataset to benchmark the performance of existing methods on the challenging motion-expression \ourdatasetICCV in \cref{tab:MeViS2023} and the newly introduced \ourdatasetTPAMI, which further include no-target and motion reasoning scenarios, in \cref{tab:MeViS2024}. We evaluate 1 modified image-based method VLT~\cite{vltpami} and 5 recent state-of-the-art video-based methods, including URVOS~\cite{seo2020urvos}, LBDT~\cite{Ding_2022_CVPR}, MTTR~\cite{MTTR}, ReferFormer~\cite{wu2022referformer}, and DsHmp~\cite{DsHmp} on the validation set of \ourdataset. 

\myparagraph{\ourdatasetICCV.} The evaluation results presented in \cref{tab:MeViS2023} show that previous state-of-the-art methods could only achieve performance ranging from \textbf{27.8\% $\mathcal{J\&F}$} to \textbf{31.0\% $\mathcal{J\&F}$} on the validation set of \ourdatasetICCV, while their results on other conventional datasets like Refer-YouTube-VOS~\cite{seo2020urvos} and DAVIS$_{17}$-RVOS~\cite{khoreva2018video} are usually above \textbf{60\% $\mathcal{J\&F}$} (see \cref{tab:ytvos_davis}).

\begin{table}[t]
   \renewcommand\arraystretch{1.16}
   \centering
   \footnotesize
   \caption{\ourdatasetTPAMI Benchmark Results. $\mathcal{J\&F}_s$ and $\mathcal{J\&F}_m$ show performance for single-target and multi-target samples, respectively. N-acc. and T-acc. represent no-target performance.}
  \vspace{-2mm}
   \setlength{\tabcolsep}{1.16mm}
   \begin{tabular}{l|ccc|cccc}
         \specialrule{.1em}{.05em}{.05em}
          \rowcolor[gray]{.9}
\rowcolor[gray]{.9} 
        \rowcolor[gray]{.9} Methods& $\mathcal{J\&F}$ & $\mathcal{J}$ & $\mathcal{F}$ &$\mathcal{J\&F}_s$&$\mathcal{J\&F}_m$&N-acc. & T-acc.\\
         \hline\hline
         URVOS~\cite{seo2020urvos}& 23.4&20.3&26.5&25.7&22.1&\ \ 0.0&100.0\\
         LBDT~\cite{Ding_2022_CVPR}&25.1&21.8&28.4&27.5&24.3&\ \ 0.0&100.0\\
         MTTR~\cite{MTTR}&25.9&22.5&29.3&28.3&25.2&\ \ 0.0&100.0\\
         ReferFormer~\cite{wu2022referformer}&26.7&23.1&30.3&28.9&25.8&\ \ 0.0&100.0\\
         VLT+TC~\cite{vltpami}&30.1 & 26.5&33.7&31.9&32.5&\ \ 0.0&100.0\\
         DsHmp~\cite{DsHmp}&40.8&36.9&44.7&39.7&49.4&  21.0&\ \ 84.1\\
         \hline         
         \textbf{\ourmodel} (ours) &38.3 & 35.6& 40.9&37.1&46.7&20.1 &\ \ 72.9 \\
          \textbf{\ournewmodel} (ours)&\textbf{43.9}&\textbf{40.8}&\textbf{47.0}&\textbf{41.2}&\textbf{51.6}&\textbf{45.7}&\ \ \textbf{87.4}\\
         \specialrule{.1em}{.05em}{.05em}
      \end{tabular}
   \label{tab:MeViS2024}%
\end{table}%

\myparagraph{\ourdatasetTPAMI.} The evaluation results presented in \cref{tab:MeViS2024} show that previous state-of-the-art methods struggle with no-target and motion reasoning scenarios.
In contrast, the proposed LMPM++ effectively addresses these challenges by leveraging the capabilities of an embedded large language model. We further decompose the overall performance into different expression types: single-target, multi-target, and no-target cases, corresponding to $\mathcal{J\&F}_s$, $\mathcal{J\&F}_m$, and N-acc.~\&~T-acc.~in \cref{tab:MeViS2024}, respectively. The results show that multi-target and no-target samples present greater challenges for methods that only output the top-1 object mask, such as ReferFormer~\cite{wu2022referformer}. This limitation arises because the top-1 strategy assumes the presence of a single target object, a default assumption in previous RVOS datasets such as Refer-YouTube-VOS\cite{seo2020urvos} and DAVIS$_{17}$-RVOS~\cite{khoreva2018video}. As a result, it is inherently incapable of handling no-target samples, leading to very low (even 0) N-acc.~scores. For multi-target cases, even if the top-1 mask is highly accurate, it can only capture one of the target objects. In contrast, methods using adaptive output strategies, such as DsHmp~\cite{DsHmp} and the proposed LMPM++, perform much better in multi-target scenarios. In multi-target scenarios, partially erroneous predictions have a smaller impact on performance due to the larger ground truth area, whereas an error in single-target cases leads to a significantly lower score due to small target area. These results highlight that the proposed \ourdatasetTPAMI dataset presents significant challenges for evaluating models’ generalization abilities across a variety of complex scenarios.

Our experiments on \ourdatasetICCV and \ourdatasetTPAMI show that while notable progress has been made in language-guided video object segmentation on existing benchmarks, the new challenges introduced by \ourdataset underline the need for further exploration of motion expression-guided video segmentation in complex scenarios. These challenges can arise from various factors across both linguistic and visual modalities, such as the use of motion expressions, highly dynamic objects, and fast-paced motions in videos, all of which can adversely affect overall performance. 
Besides, \ourdatasetICCV primarily focuses on explicitly referring to single or multiple objects, neglecting real-world scenarios where expressions may involve implicit information or might be intentionally or unintentionally incorrect, resulting in no objects being referred to. In \ourdatasetTPAMI, we introduce no-target expressions and motion reasoning expressions to make the dataset more reflective of real-world situations. Consequently, \ourdatasetTPAMI is more challenging than \ourdatasetICCV, as evidenced by DsHmp's~\cite{DsHmp} $\mathcal{J\&F}$ score being 5.6\% (40.8\% \vs 46.4\%) lower on \ourdatasetTPAMI compared to \ourdatasetICCV. \ourdatasetTPAMI offers a dataset that is more representative of real-world complexities and closer to practical applications like embodied AI.

\begin{figure}[t]
    \centering
    \includegraphics[width=\linewidth]{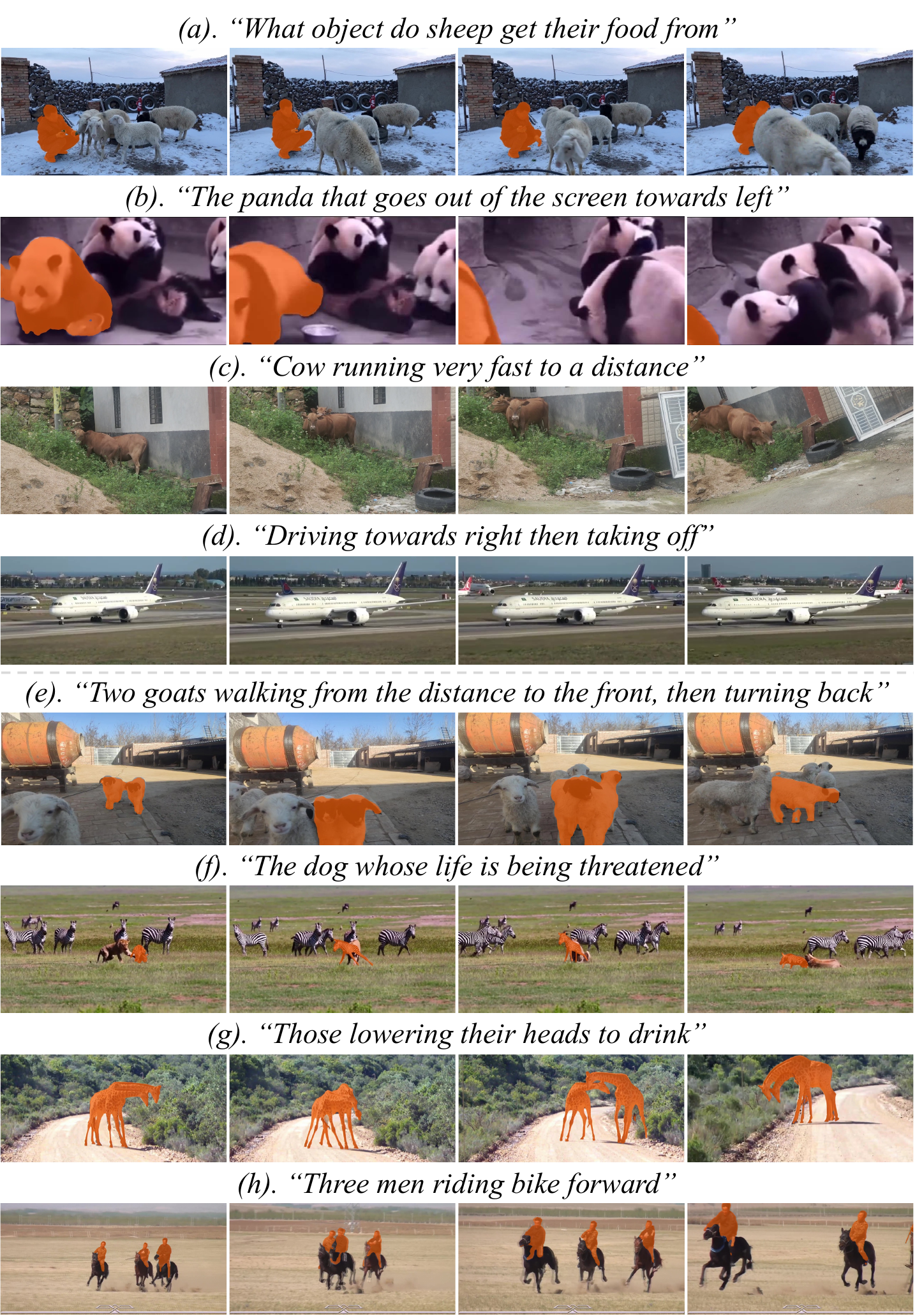}
    \vspace{-6.6mm}
    \caption{Examples (a)-(d) show success cases, while (e)-(h) display failure cases of \ournewmodel. Examples (c)-(d) and (f)-(h) correspond to no-target samples.}
    \label{fig:visualizations}
\end{figure}

\myparagraph{Visualizations.} 
\cref{fig:visualizations} presents both successful and unsuccessful cases using the proposed approach \ournewmodel. Examples (a) and (b) show successful cases where \ournewmodel accurately interprets expressions requiring reasoning and long-term motion tracking, such as ``\textit{get their food}" and ``\textit{goes out of the screen}.'' {Examples (c) and (d) demonstrate successful cases for no-target expressions. For the expression ``\textit{Cow running very fast to a distance}'' in (c), LMPM++ correctly avoids outputting a mask, indicating its ability to understand concepts like ``\textit{fast}'' and ``\textit{distance}'' and recognize their mismatch with the visual content.
In (d), LMPM++ accurately perceives directional cues like ``\textit{right}'', showcasing its spatial understanding.} {On the other hand, examples (e) to (h) present failure cases.}
Example (e) involves a sentence describing a long-term motion with multiple target objects. While our method initially identifies the correct targets, \ie, the ``\textit{two goats walking from the distance}'', the targets are lost in the later stages of the video when the object motions become more complex and intertwined. {(f)-(h) highlight the inherent difficulty of no-target scenarios, particularly when dealing with fine-grained actions, complex object relationships, and subtle contextual differences.}
In example (f), the expression contains misleading language, where the entity in danger, chased by a lion, should be identified as a ``\textit{zebra}'' instead of a ``\textit{dog}''. Our method fails to disambiguate this confusing term, leading to an incorrect prediction. {In (g), the model fails when the described action is highly similar to the action in the video, such as distinguishing between ``\textit{lowering their heads to eat}'' and ``\textit{lowering their heads to drink}''. In (h), challenges arise when the motion involves complex interactions with other objects.}
These failure cases underscore the challenges posed by \ourdataset, highlighting the need for models to possess a strong understanding of motion expression, world knowledge for effective reasoning, and an unbiased approach to scenarios where no target may be present.

\subsection{Results on Previous RVOS Datasets}

\myparagraph{Refer-YouTube-VOS \& DAVIS$_{17}$-RVOS.} 
In \cref{tab:ytvos_davis}, we report our results on Refer-YouTube-VOS~\cite{seo2020urvos} and DAVIS$_{17}$-RVOS~\cite{khoreva2018video} datasets, where our method surpasses all existing approaches across various metrics.
For Refer-YouTube-VOS, our model \ournewmodel with the Video-Swin-Tiny backbone achieves a score of 64.0\% \( \mathcal{J}\)\&\( \mathcal{F}\), which is an improvement of 0.3\% over the previous best LoSh~\cite{LoSh}. When utilizing a larger backbone, specifically the Video-Swin-Base, our model's performance further elevates to 67.8\% \( \mathcal{J}\)\&\( \mathcal{F}\), consistently outperforming all other methods by at least 0.6\%. On DAVIS$_{17}$-RVOS dataset, \ournewmodel achieves the best performance of 65.0\% \( \mathcal{J}\)\&\( \mathcal{F}\) with Video-Swin-Base.

\begin{table}[t!]
\setlength{\tabcolsep}{3.16pt}
\centering
\footnotesize
\caption{Results on Refer-YouTube-VOS and DAVIS$_{17}$-RVOS.
} \label{tab:ytvos_davis}
\vspace{-2mm}
\begin{tabular}{r | c |c c c | c c c}
 \rowcolor[gray]{.9}
\specialrule{.1em}{.05em}{.05em}
&  &  \multicolumn{3}{c |}{Refer-YouTube-VOS} & \multicolumn{3}{c}{DAVIS$_{17}$-RVOS} \\
\rowcolor[gray]{.9} 
Method &Reference & \( \mathcal{J} \)\&\( \mathcal{F} \) & \( \mathcal{J} \) & \( \mathcal{F} \)  &  \( \mathcal{J} \)\&\( \mathcal{F} \) & \( \mathcal{J} \) & \( \mathcal{F} \) \\
\hline 
\hline
\multicolumn{8}{c}{Video-Swin-Tiny}\\ 
\hline
ReferFormer~\cite{wu2022referformer} & \pub{CVPR'22} & 59.4 & 58.0 & 60.9  & 59.6 & 56.5 & 62.7  \\
HTML~\cite{HTML} &\pub{ICCV'23} & 61.2 &59.5 &63.0&-&-&-\\
R$^2$-VOS~\cite{R2VOS}& \pub{ICCV'23} & 61.3 &59.6 &63.1 &-&-&-\\
SgMg~\cite{SgMg}  & \pub{ICCV'23} & {62.0} & {60.4} & {63.5}  & {61.9} & {59.0} & {64.8} \\ 
TempCD~\cite{TempCD}&\pub{ICCV'23} &62.3&60.5& 64.0& 62.2 &59.3& 65.0 \\
SOC~\cite{SOC} &\pub{NeurIPS'23}&62.4& 61.1& 63.7 &63.5 &60.2 &66.7\\
DsHmp~\cite{DsHmp}&\pub{CVPR'24}&{63.6} &{61.8} & {65.4} & {64.0}& {60.8}&{67.2} \\
LoSh~\cite{LoSh}&\pub{CVPR'24}&63.7 &62.0 &65.4&62.9 &60.1 &65.7 \\
\hline
\textbf{\ournewmodel} (ours)&\pub{TPAMI}&\textbf{64.0} &\textbf{62.2} &\textbf{65.8} & \textbf{64.2} & \textbf{60.9}& \textbf{67.5} \\
\hline
\multicolumn{8}{c}{Video-Swin-Base}\\
\hline
ReferFormer~\cite{wu2022referformer} & \pub{CVPR'22} & 62.9 & 61.3 & 64.6 & 61.1 & 58.1 & 64.1 \\
OnlineRefer~\cite{OnlineRefer} & \pub{ICCV'23} &62.9& 61.0 &64.7 &62.4 &59.1 &65.6\\
HTML~\cite{HTML} &\pub{ICCV'23} & 63.4 &61.5 &65.2 &62.1 &59.2 &65.1\\
SgMg~\cite{SgMg}  & \pub{ICCV'23} &  {65.7} & {63.9} & {67.4}  & {63.3} & {60.6} & {66.0} \\ 
TempCD~\cite{TempCD}&\pub{ICCV'23} &65.8&63.6 &68.0 & 64.6& 61.6& 67.6\\
SOC~\cite{SOC} & \pub{NeurIPS'23} &66.0 &64.1& 67.9& 64.2 &61.0 &67.4\\
DsHmp~\cite{DsHmp}&\pub{CVPR'24}&{67.1}& {65.0}& {69.1}&{64.9} &{61.7} &{68.1}\\
LoSh~\cite{LoSh}&\pub{CVPR'24}&67.2 &65.4& 69.0 &64.3 &61.8 &66.8\\
\hline
\textbf{\ournewmodel} (ours)&\pub{TPAMI}&\textbf{67.8}& \textbf{65.7}& \textbf{69.9}&\textbf{65.0} &\textbf{61.9} &\textbf{68.1}\\
\specialrule{.1em}{.05em}{.05em}
\end{tabular}
\end{table}

\myparagraph{A2D Sentence \& J-HMDB Sentence.} 
We further evaluate the performance of \ournewmodel on A2D Sentence and J-HMDB Sentence datasets~\cite{gavrilyuk2018actor}, as shown in \cref{tab:A2D_JHMDB}. Consistent with the approach in~\cite{wu2022referformer}, the models are initially pre-trained on RefCOCO/+/g and subsequently fine-tuned on A2D Sentence. The J-HMDB Sentence dataset is only used for evaluation purposes.
The proposed \ournewmodel achieves new state-of-the-art results, surpassing the closest competitor, DsHmp~\cite{DsHmp}, by 0.9\% mAP on A2D Sentence and 0.4\% mAP on J-HMDB Sentence, respectively.

The performance gains of \ournewmodel on the above four datasets, while notable, are relatively modest when compared to those on MeViS. This could be attributed to the nature of these four datasets, which primarily contain sentences with image-level descriptions for the first frame and do not strictly necessitate motion expressions. Nevertheless, \ournewmodel maintains its state-of-the-art status, demonstrating its broad applicability and effectiveness.

\begin{table}[t!]
\footnotesize
\centering
\setlength{\tabcolsep}{3.16pt}
\caption{Results on A2D Sentence and J-HMDB Sentence.} \label{tab:A2D_JHMDB}
\vspace{-2mm}
\begin{tabular}{r | c | c c c | c c c }
 \rowcolor[gray]{.9}
\specialrule{.1em}{.05em}{.05em}
 && \multicolumn{3}{c |}{A2D Sentence	} & \multicolumn{3}{c}{J-HMDB Sentence} \\\rowcolor[gray]{.9}
 Method& Reference& mAP &oIoU & mIoU & mAP & oIoU & mIoU \\
\hline
\hline
\multicolumn{8}{c}{Video-Swin-Tiny}\\
\hline
MTTR~\cite{MTTR} &\pub{CVPR'22}& 46.1 & 72.0 & 64.0 & 39.2 & 70.1 & 69.8  \\
ReferFormer~\cite{wu2022referformer} &\pub{CVPR'22}&  52.8 & 77.6 & 69.6 & 42.2 & 71.9 & 71.0  \\ 
HTML~\cite{HTML}&\pub{ICCV'23}&53.4 & 77.6 &69.2 &42.7&-&-   \\ 
SOC~\cite{SOC}&\pub{NeurIPS'23}&54.8 &78.3& 70.6 &42.7 &72.7 &71.6\\
SgMg~\cite{SgMg} &\pub{ICCV'23}& {56.1} & {78.0} & {70.4} & {44.4} & {72.8} & {71.7} \\ 
{DsHmp}~\cite{DsHmp}&\pub{CVPR'24}&{57.2}&{79.0}&{71.3}&{44.9}&{73.1}&{72.1} \\
LoSh~\cite{LoSh}&\pub{CVPR'24}&57.6&79.3& 71.6&-&-&-\\
\hline
\textbf{\ournewmodel} (ours)&\pub{TPAMI}&\textbf{57.8}&\textbf{79.4}&\textbf{71.6}&\textbf{45.3}&\textbf{73.4}&\textbf{72.4} \\

\hline
\multicolumn{8}{c}{Video-Swin-Base}\\
\hline
ReferFormer~\cite{wu2022referformer} &\pub{CVPR'22}&  55.0 & 78.6 & 70.3 & 43.7 & 73.0 & 71.8\\
OnlineRefer~\cite{OnlineRefer} &\pub{ICCV'23}& -&79.6& 70.5&- &73.5 &71.9\\

HTML~\cite{HTML}&\pub{ICCV'23}& 56.7 & 79.5 &71.2 &44.2&-&-\\ 

SOC~\cite{SOC}&\pub{NeurIPS'23}&57.3  &80.7 & 72.5 & 44.6 &73.6 &72.3\\
SgMg~\cite{SgMg} &\pub{ICCV'23}&  {58.5} & {79.9} & {72.0} & {45.0} & {73.7} & {72.5}  \\ 
DsHmp~\cite{DsHmp}&\pub{CVPR'24}&{59.8}&{81.1}&{72.9}&{45.8}&{73.9}&{73.0}\\
LoSh~\cite{LoSh}&\pub{CVPR'24}&59.9&81.2& 73.1&-&-&-\\
\hline
\textbf{\ournewmodel} (ours)&\pub{TPAMI}&\textbf{60.7}&\textbf{81.8}&\textbf{73.9}&\textbf{46.2}&\textbf{74.8}&\textbf{73.4}\\
\specialrule{.1em}{.05em}{.05em}
\end{tabular}
\end{table}

\begin{table}[t]
   \renewcommand\arraystretch{1.06}
   \centering
   \footnotesize
   \caption{Results of RMOT on \ourdatasetTPAMI. * indicates that the RMOT metrics are adapted to account for no-target samples.}
  \vspace{-2mm}
   \setlength{\tabcolsep}{1.36mm}{\begin{tabular}{l|c|ccccc}
         \specialrule{.1em}{.05em}{.05em}
        \rowcolor[gray]{.9} Method & Reference & HOTA* & DetA* & AssA* & N-acc. & T-acc. \\
         \hline\hline 
         TransRMOT \cite{rmot} & \pub{CVPR'23} & 18.6 & \ \ 9.2 & 38.1 & \textbf{56.9} & 52.3 \\
         TempRMOT \cite{zhang2024bootstrapping_rmot} & \pub{PrePrint} & 30.0 & 17.7 & {51.3} & 18.2 & 72.3 \\ 
         \hline
         \textbf{\ournewmodel} & \pub{TPAMI} & {38.1} & {28.1} & {52.5} & 45.7 &  \textbf{87.4}\\
         {\textbf{LMPM++}$_{\rm{det}}$}  & \pub{TPAMI} & {\textbf{38.8}} & {\textbf{29.0}} & {\textbf{52.7}} & {45.7}&  {\textbf{87.4}}\\
         \specialrule{.1em}{.05em}{.05em}
      \end{tabular}}%
   \label{tab:RMOT}%
\end{table}%

\begin{table}[t]
   \renewcommand\arraystretch{1.16}
   \centering
   \footnotesize
   \caption{Results of AVOS on \ourdatasetTPAMI.}
  \vspace{-2mm}
   \setlength{\tabcolsep}{1.56mm}{\begin{tabular}{l|c|ccccc}
         \specialrule{.1em}{.05em}{.05em}
        \rowcolor[gray]{.9} Methods& Reference & $\mathcal{J\&F}$ & $\mathcal{J}$ & $\mathcal{F}$ & N-acc. & T-acc. \\
         \hline\hline 
          WNet \cite{pan2022wnet} & \pub{CVPR'22} & 16.5 & 16.6& 16.3 & \ \ 0.5 & \ \ 99.3\\
          MUTR \cite{mutr} & \pub{AAAI'24} & 33.6 & 30.9 & 36.3 & \ \ 0.0 & 100.0 \\
          \hline
          LMPM\textsubscript{Audio$\rightarrow$Text} \cite{MeViS} & \pub{ICCV'23} & 36.8 & 33.2 & 40.5 & 22.1 & \ \ 64.7 \\
          LMPM\textsubscript{Audio} \cite{MeViS} & \pub{ICCV'23} & 37.5 & 34.5 & 40.4 & 11.0 & \ \ 78.8 \\
          LMPM\textsubscript{GT Text} \cite{MeViS} & \pub{ICCV'23} & 38.3 & 35.6 & 40.9 & 20.1 & \ \ 72.9\\
          \hline
          \textbf{\ournewmodel} (ours) & \pub{TPAMI} & \textbf{42.3} &\textbf{39.1}& \textbf{45.5} & \textbf{43.2} & \ \ \textbf{85.4} \\
         \specialrule{.1em}{.05em}{.05em}
      \end{tabular}}%
   \label{tab:avos}%
\end{table}%

\subsection{{Referring Multi-Object Tracking Results}}\label{sec:eval_RMOT}

As mentioned in \cref{sec:supportedTasks}, \ourdatasetTPAMI supports the RMOT task. \cref{tab:RMOT} presents RMOT results on \ourdatasetTPAMI, comparing \ournewmodel with previous state-of-the-art methods \cite{rmot, zhang2024bootstrapping_rmot}. {For \ournewmodel, bounding boxes are derived from the bounding rectangles of the masks, while LMPM++$_{\rm{det}}$ integrates an additional detection head}. The metrics HOTA*, DetA*, and AssA* are adapted from RMOT \cite{rmot} to account for no-target samples in \ourdatasetTPAMI. Originally, HOTA scored 0 for no-target samples; we modify this so that a score of 1 is awarded when the model accurately predicts an empty bounding box trajectory, otherwise scoring 0.
N-acc.~and T-acc.~\cite{GRES} are also employed to evaluate performance on no-target identification. 
As shown in \cref{tab:RMOT}, the proposed \ournewmodel outperforms other methods with the highest performance across several metrics, including HOTA* (38.1\%), DetA* (28.1\%), and T-acc. (87.4\%). While TransRMOT \cite{rmot} slightly outperforms in N-acc. (56.9\% vs. 45.7\%), our model shows significant overall improvements, especially in detection and target prediction accuracy. {Compared to \ournewmodel, LMPM++$_{\rm{det}}$ achieves better performance with an additional detection head for direct bounding box prediction and supervision using ground truth annotations.}
The above indicates the superior transfer performance of our method among different tasks.

\subsection{Audio-Guided Video Object Segmentation}
\cref{tab:avos} benchmarks AVOS methods on \ourdatasetTPAMI dataset. 
WNet \cite{pan2022wnet} and MUTR \cite{mutr} are models that originally support audio as input, but they only achieve \( \mathcal{J}\)\&\( \mathcal{F}\) scores of 16.5\% and 33.6\%, respectively, highlighting the difficulty of MeViS. MUTR's N-acc.~of 0\% and T-acc.~of 100\% indicate that the inclusion of no-target cases significantly increases the challenge of the MeViS dataset, especially for models that tend to output one target for any given expression. 
For LMPM \cite{MeViS}, three variants are tested: LMPM\textsubscript{GT Text}, LMPM\textsubscript{Audio$\rightarrow$Text}, and LMPM\textsubscript{Audio}. LMPM\textsubscript{GT Text} uses ground truth text as input, achieving the highest \( \mathcal{J}\)\&\( \mathcal{F}\) score and highlighting the advantage of accurate text data. LMPM\textsubscript{Audio$\rightarrow$Text} converts audio to text using the Whisper-Base \cite{whisper} model before inputting it into LMPM, but it slightly underperforms due to transcription accuracy issues. Finally, LMPM\textsubscript{Audio} directly extracts audio features using Whisper, replacing RoBERTa\cite{liu2019roberta}, but this approach faces additional challenges and lags behind the text-based variant, reflecting the complexities of direct audio processing.
\ournewmodel achieves the highest \( \mathcal{J}\)\&\( \mathcal{F}\) score of 42.3\%, outperforming all LMPM variants and other methods on the AVOS task. It excels in both N-acc.~(43.2\%) and T-acc.~(85.4\%), demonstrating superior robustness and accuracy, particularly in handling no-target cases. This highlights the significant improvements \ournewmodel brings over previous approaches.

\begin{table}[t]
   \renewcommand\arraystretch{1.16}
   \centering
   \footnotesize
   \caption{Results of RMEG on \ourdataset.}
  \vspace{-2mm}
   \setlength{\tabcolsep}{3.96mm}{\begin{tabular}{l|c|ccc}
         \specialrule{.1em}{.05em}{.05em}
        \rowcolor[gray]{.9} Methods& Reference  & METEOR & CIDEr \\
         \hline\hline 
         GIT~\cite{wang2022git} & \pub{TMLR'22}& 12.33& 18.20\\
         VAST~\cite{chen2024vast} & \pub{NeurIPS'23} &{10.66}& {20.42}\\
         NarrativeBridge~\cite{nadeem2024narrativebridge} & \pub{PrePrint}&{14.99}& {25.68}\\
         VideoLLaMA~2~\cite{cheng2024videollama2advancingspatialtemporal} & \pub{PrePrint} &{15.68}& {27.10}\\
         \specialrule{.1em}{.05em}{.05em}
      \end{tabular}}%
   \label{tab:rmeg}%
   \vspace{-2mm}
\end{table}%
\subsection{{Referring Motion Expression Generation Results}}\label{sec:eval_RMEG}

Herein we benchmark several existing methods on the proposed task of Referring Motion Expression Generation (RMEG). Following existing works in classic video and image captioning tasks, we employ two commonly used metrics to evaluate model performance: METEOR~\cite{denkowski2014meteor} and CIDEr~\cite{vedantam2015cider}. As shown in \cref{tab:rmeg}, we benchmark four methods on MeViS, including two traditional video captioning methods GIT~\cite{wang2022git} and VAST~\cite{chen2024vast}, one video captioning model that involves Large Language Models (LLM), NarrativeBridge~\cite{nadeem2024narrativebridge}, and one native LLM VideoLLaMA2~\cite{cheng2024videollama2advancingspatialtemporal}. As most of the traditional methods do not support specifying certain objects, we highlight the target object with a semitransparent mask throughout the video to help the methods focus on the object. Additionally, we instruct the LLM-based methods with the language prompt:  \textit{``Write an unambiguous referring expression that unambiguously describes the motion of the masked object(s) in the video.''}

From \cref{tab:rmeg}, the METEOR scores of most methods are less than 15, indicating that the generated motion expressions lack accuracy. The CIDEr scores are also low, with the highest score being 27.10. 
We find that the LLM-based methods, such as NarrativeBridge~\cite{nadeem2024narrativebridge} and VideoLLaMA~2~\cite{cheng2024videollama2advancingspatialtemporal}, outperform traditional methods by a significant margin, \eg, NarrativeBridge~\cite{nadeem2024narrativebridge} exceeds VAST~\cite{chen2024vast} by over 5 points on the CIDEr metric. This suggests that the RMEG task requires strong reasoning abilities to unambiguously describe the motion of the target object in the video, which LLM-based methods may handle more effectively.

\begin{figure}
   \centering
   \includegraphics[width=0.96\linewidth]{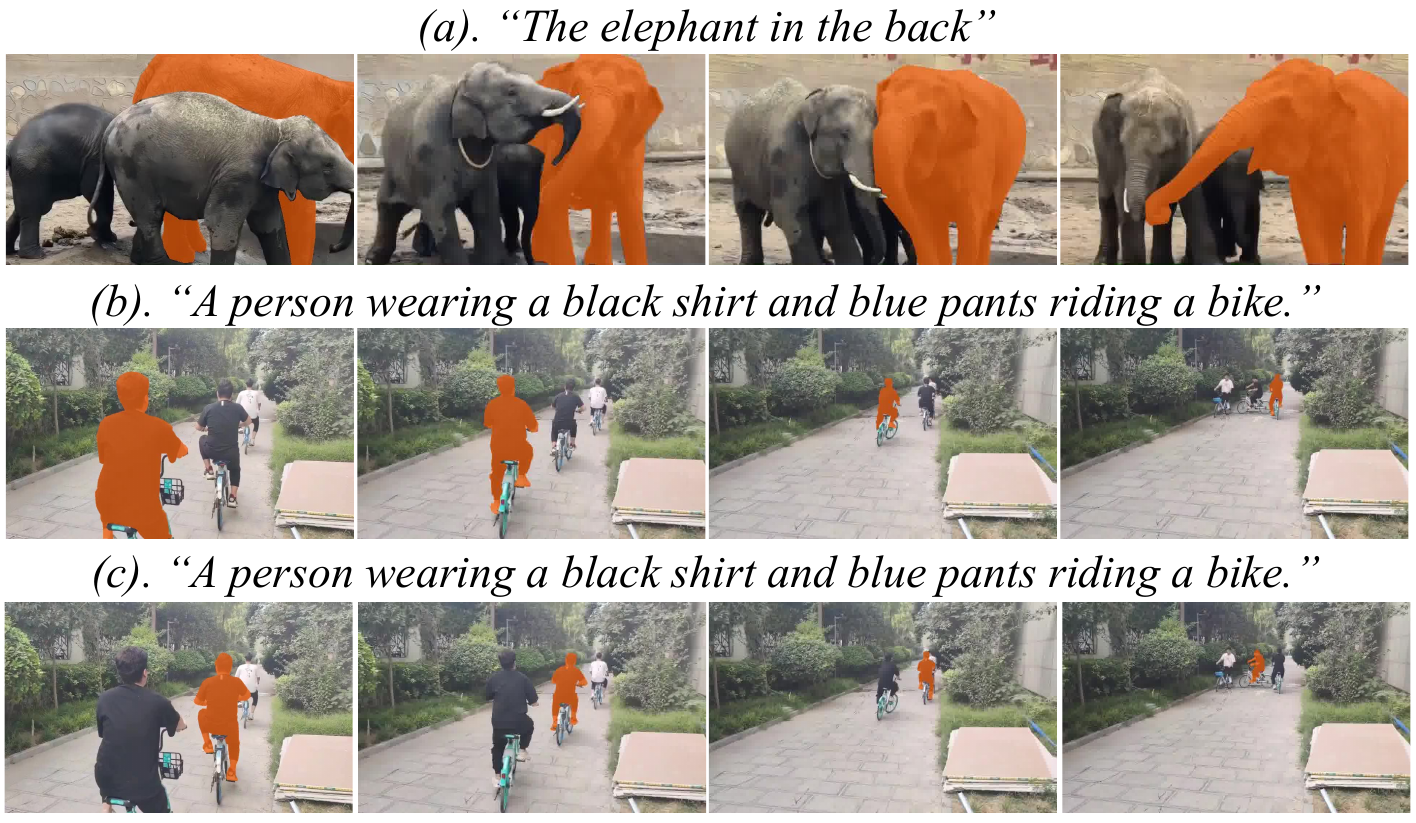}
   \vspace{-2mm}
   \caption{{REMG failure cases of VideoLLaMA 2~\cite{cheng2024videollama2advancingspatialtemporal} on MeViS.}}
   \label{fig:rmeg_fail}
   \vspace{-2mm}
\end{figure}

\cref{fig:rmeg_fail} shows some failure cases of VideoLLaMA2~\cite{cheng2024videollama2advancingspatialtemporal}. We observe two significant drawbacks of previous video captioning methods on the RMEG task. 
First, the generated expression may fail to describe the motion and only output the object state in a certain time range. For example, in \cref{fig:rmeg_fail}~(a), the predicted expression, \textit{``The elephant in the back''}, is a non-motion description and only valid for the target at the beginning of the video.
Second, when multiple similar objects are present in the video, the generated expressions may fail to distinguish between them or may even produce identical expressions for different objects. This loss of the ``referring” property is unacceptable for RMEG. For example, in \cref{fig:rmeg_fail}~(b) and (c), two people with similar appearances and movement trajectories are located in different places. The model fails to distinguish between them and generates the same expression for both, which is inadequate for the RMEG task.

\section{Conclusion and Discussion}

This paper introduces a large-scale multi-modal dataset \ourdatasetTPAMI, designed to advance research in referring video understanding, especially with motion-centric language descriptions across diverse and complex scenarios. Through extensive benchmarks on \ourdataset, we demonstrate the limitations of existing methods in RVOS, AVOS, RMOT, and RMEG tasks, showing that these methods fall short in effectively leveraging motion expressions for video understanding. Moreover, we analyze the challenges and propose a baseline approach \ournewmodel to meet the challenges of the proposed \ourdataset dataset.

\textbf{Future Directions.} 
There are many interesting research directions and remaining challenges to be addressed with the \ourdataset dataset. 
These include but are not limited to: (\romannumeral1) developing techniques for enhanced motion understanding and representation in both visual and linguistic domains, (\romannumeral2) designing more rigorous and robust models that can effectively handle diverse motion types spanning across a range of frames, including long-term, short-term, and complex motions, (\romannumeral3) developing advanced models that can handle complex scenes with various types of objects and expressions, (\romannumeral4) creating more efficient models that can effectively reduce the number of detected redundant objects, (\romannumeral5) designing effective cross-modal fusion methods to better align the information between language and visual signals, (\romannumeral6) investigating the potential of transfer learning and domain adaptation in language-guided video segmentation, (\romannumeral7) developing methods that can better handle the open-world concepts in both the visual and linguistic domain. These challenges require significant research efforts to advance the research in language-guided video understanding.

{
\bibliographystyle{IEEEtran}
\bibliography{egbib}
}
\end{document}